\documentclass[journal]{IEEEtran}

\usepackage{amsmath,amsfonts}
\usepackage[ruled,linesnumbered]{algorithm2e}
\usepackage{array}
\usepackage{subfigure}
\usepackage{textcomp}
\usepackage{stfloats}
\usepackage{url}
\usepackage{verbatim}
\usepackage{graphicx}
\usepackage{cite}
\usepackage{multirow}
\usepackage[square,comma,numbers,sort&compress]{natbib}
\usepackage{booktabs}
\usepackage{caption}
\usepackage{color}

\begin{document}

\title{Efficient Bayesian Policy Reuse with a Scalable Observation Model in Deep Reinforcement Learning}

%
%

\author{Jinmei Liu,~\IEEEmembership{Graduate Student Member,~IEEE}~Zhi~Wang,~\IEEEmembership{Member,~IEEE},~Chunlin~Chen,~\IEEEmembership{Senior Member,~IEEE},~Daoyi~Dong,~\IEEEmembership{Fellow,~IEEE}
\thanks{This work is published on \textit{IEEE Transactions on Neural Networks and Learning Systems}, 2023, DOI: 10.1109/TNNLS.2023.3281604.}
\thanks{The work was supported in part by the National Natural Science Foundation of China under Grant 62006111 and Grant 62073160, in part by the Australian Research Council's Future Fellowship funding scheme under Project FT220100656, and in part by the Natural Science Foundation of Jiangsu Province of China under Grant BK20200330. \textit{(Corresponding authors: Zhi Wang, Chunlin Chen)}}
\thanks{J. Liu, Z. Wang, and C. Chen are with the Department of Control Science and Intelligent Engineering, School of Management and Engineering, and the Research Center for Novel Technology of Intelligent Equipment, Nanjing University, Nanjing 210093, China. (email: jinmeiliu@smail.nju.edu.cn, \{zhiwang, clchen\}@nju.edu.cn).}
\thanks{D. Dong is with the School of Engineering and Information Technology, University of New South Wales, Canberra, ACT 2600, Australia (email: daoyidong@gmail.com).}
}

\maketitle

\begin{abstract} 
Bayesian policy reuse (BPR) is a general policy transfer framework for selecting a source policy from an offline library by inferring the task belief based on some observation signals and a trained observation model.
In this paper, we propose an improved BPR method to achieve more efficient policy transfer in deep reinforcement learning (DRL).
First, most BPR algorithms use the episodic return as the observation signal that contains limited information and cannot be obtained until the end of an episode.
Instead, we employ the state transition sample, which is informative and instantaneous, as the observation signal for faster and more accurate task inference.
Second, BPR algorithms usually require numerous samples to estimate the probability distribution of the tabular-based observation model, which may be expensive and even infeasible to learn and maintain, especially when using the state transition sample as the signal.
Hence, we propose a scalable observation model based on fitting state transition functions of source tasks from only a small number of samples, which can generalize to any signals observed in the target task.
Moreover, we extend the offline-mode BPR to the continual learning setting by expanding the scalable observation model in a plug-and-play fashion, which can avoid negative transfer when faced with new unknown tasks.
Experimental results show that our method can consistently facilitate faster and more efficient policy transfer.

\end{abstract}


\section{Introduction}

Reinforcement learning (RL)~\cite{sutton2018reinforcement} is a general optimization framework of how an artificial agent learns an optimal policy to maximize the cumulative reward by interacting with its environment.
RL has been used to find optimal controllers ~\cite{ mazouchi2021data, yang2021model, yang2021hamiltonian} and realize human-computer interaction~\cite{chavarriaga2007err}, and has a broad prospect in other practical applications~\cite{wang20192}.
With recent advances, deep RL (DRL) has achieved state-of-the-art performance on various tasks~\cite{li2020deep,wang2022lifelong,wei2021deep}, such as video games~\cite{vinyals2019grandmaster}, board games~\cite{silver2017mastering}, robotics~\cite{hwangbo2019learning}, autonomous driving~\cite{huang2022efficient}, and quantum information~\cite{li2020quantum}.
However, DRL algorithms are usually sensitive to the choice of hyper-parameters~\cite{henderson2018deep,Majumdar20evolu} and typically require numerous samples to converge to good policies~\cite{yarats2019improving,Mitchell21offline}, which can be computationally intensive in real-world applications.

To address this concern, transfer RL~\cite{taylor2009transfer} is usually introduced to reduce the number of samples required for learning a target task by reusing previously acquired knowledge from relevant source tasks~\cite{xu20knowledge,abdolshah21new,multisource}.
Reusing policies are widely investigated kind of approaches for this topic~\cite{gupta2017learning,parisotto2015actor,yang2020single,barreto2016successor,brys2015policy,yu2021protective}, since they are intuitive, direct, and do not rely on value functions that may be difficult or unavailable to transfer.

In this paper, we focus on the problem of transferring policies from multiple source tasks in DRL.
Bayesian policy reuse (BPR)~\cite{rosman2016bayesian} is a general transfer framework for responding to a target task by selecting the most appropriate policy from an offline library based on some observation signals and a trained observation model.
Moreover, BPR has been successfully applied to handle non-stationary opponents in multi-agent systems~\cite{hernandez2016bayesian,yang2018towards,zheng18deep,gao2022bayesian}.
However, BPR algorithms have several limitations: 1) they usually use the episodic return as the observation signal that is only a scalar containing limited information, and the learning can be too slow as they must wait until the end of a full episode before updating the task belief;
2) they typically require numerous samples to estimate the probability distribution of the tabular-based observation model, which could be expensive and even infeasible to learn and maintain, especially using the state transition sample as the signal; and 3) they need to apply a fixed set of policies on all source tasks to obtain the tabular observation model in an offline manner, impeding the applicability to real-world continual learning settings.


We address the above limitations in the paper.
First, we employ the state transition sample $(s,a,r,s')$, the four-tuple composed of state, action, reward, and the next state, as the observation signal that reveals more task information since the state transition function $\mathcal{P}(s',r|s,a)$ completely characterizes the task's dynamics.
Then, we use only a small number of samples to fit the state transition functions of source tasks, which are utilized to compute the scalable observation model that can generalize to any signals observed in the target task.
In particular, we adopt two typical approaches, the non-parametric Gaussian process (GP) and the parametric neural network (NN), to estimate these functions, while in principle any distribution matching technique or probabilistic model can be adopted for this estimation. 
GP exhibits better sample efficiency and uncertainty measurements on the predictions, while NN can scale to extremely high-dimensional tasks.
Finally, although there are several BPR+ algorithms~\cite{hernandez2016bayesian,yang2018towards,zheng18deep,zheng2021efficient,gao2022bayesian} that can incorporate new models online, they still need to retrain the observation model offline when adding a new source policy into the library.
Instead, we extend the offline-mode BPR to continual learning settings in a truly online manner by expanding the scalable observation model in a plug-and-play fashion when incorporating new source policies. Experimental results show that our method achieves more efficient policy reuse, and realizes effective continual learning with avoiding negative transfer.



In summary, our main contributions include: i) employing the state transition sample as the observation signal and proposing a scalable observation model to achieve efficient policy reuse in DRL compared to the basic BPR algorithms~\cite{rosman2016bayesian,hernandez2016bayesian,yang2018towards,zheng18deep,gao2022bayesian,zheng2021efficient}; and ii) extending the offline BPR to continual learning in an online manner compared to BPR+ algorithms~\cite{hernandez2016bayesian,yang2018towards,zheng18deep,zheng2021efficient,gao2022bayesian}, which avoids negative transfer and realizes better applicability to real-world applications.

The remainder of the paper is organized as follows. 
Section~\ref{background} introduces the background including preliminaries of BPR and the related work.
Section~\ref{method} presents our method in detail.
Section~\ref{experiments} shows the experimentation, and Section~\ref{conclusions} presents concluding remarks and future work.

\section{Background}\label{background}
\subsection{Preliminaries of BPR}\label{preliminiaries}
BPR provides an efficient framework for an agent to perform well by selecting the most appropriate policy from the policy library $\Pi$ to reuse when facing a target task. 
Formally, a task $\tau \in \mathcal{T}$ is defined as a Markov Decision Process (MPD), and a policy $\pi \in \Pi$ outputs an appropriate action $a$ given the state $s$.
Then, the return is defined as the accumulated discounted reward $U=\sum_{i=1}^{M} \gamma^{i} r_{i}$, which is received from interacting with the environment under the guidance of policy $\pi$ over an episode of M steps, where $r_{i}$ is the instantaneous reward, and $\gamma$ is the discount factor. 
BPR uses an observation model $\mathrm{P}(\sigma|\tau, \pi)$, which is a probability distribution over the observation signal $\sigma \in \Sigma$, to describe possible results when policy $\pi$ behaves on task $\tau$. 
The belief model $\beta(\mathcal{T})$, which is a probability distribution over $\mathcal{T}$, describes the similarity between the target task  $\tau_0$ and the source tasks $\mathcal{T}$.
BPR initializes the belief model $\beta^{0}(\mathcal{T})$ with a prior probability and updates it based on the observation model and the observation signal by using Bayes' rule:
\begin{equation}
\label{bpr}
\begin{aligned}
\beta^{t}(\tau) &=\frac{\mathrm{P}\left(\sigma^{t}|\tau, \pi^{t}\right) \beta^{t-1}(\tau)}{\sum_{\tau^{\prime} \in \mathcal{T}} \mathrm{P}\left(\sigma^{t} |\tau^{\prime}, \pi^{t}\right) \beta^{t-1}\left(\tau^{\prime}\right)}.
\end{aligned}
\end{equation}
To trade-off between exploration and exploitation, BPR uses the probability of expected improvement based on the task belief to select policies.
It reuses the most potential policy $\hat{\pi}$ in the policy library $\Pi$ by maximizing the expected utility as
\begin{equation}
\begin{aligned}
\hat{\pi} &=\arg \max _{\pi \in \Pi} \sum_{\tau \in \mathcal{T}} \beta(\tau) \int_{U \in R} U \mathrm{P}(U|\tau, \pi) \mathrm{d} U,
\end{aligned}
\end{equation}
where $\mathrm{P}(U|\tau, \pi)$ is the performance model, a probability distribution over the utility $U$ that describes how policy $\pi$ behaves on the task $\tau$. The notations used in the paper are summarized in Table~\ref{notation}.

\begin{table}[tb]
\setlength{\tabcolsep}{4mm}
	\centering
    \small
	\begin{tabular}{cc}
		\toprule
		Notation		& Description    \\
		\midrule	
	    $s$	    &  State     \\
  		$a$	    &  Action     \\
        $r$     &  Reward     \\
        $\mathcal{P}$ & Transition function \\
        $\pi$ & Policy \\
        $\gamma$ & Discount factor \\
        $M$ & The maximum number of steps \\
        $U$  & Episode total discounted reward function \\
        $K$ & Number of episodes \\
        $\tau$ & Task instance \\
        $\mathcal{T}$ & Task set \\
        $\Pi$  &  Policy library \\
        $\sigma$ & Observation signal \\
        $\beta$ & Belief model \\
        $\mathrm{P}$ & Probability distribution \\
        $\mathcal{D}$ &  State transition sample dataset \\
        $\vartheta$  & Neural network parameters \\
        $\mathcal{N}$ & Gaussian function \\
        $\kappa$ & Kernel function \\ 
        $\delta$ &  Signal variance of the RBF \\
        $l$      &  Characteristic length scale of the RBF \\
	   \bottomrule
	\end{tabular}
	\caption{Notations used in the paper.}
	\label{notation}
    \vspace{-0.8cm}
    
\end{table}

\subsection{Related Work}\label{related_work}
Knowledge transfer has received increasing attention recently and a wide variety of methods have been studied in the RL community~\cite{taylor2009transfer,tao21repaint}.
Lazaric et al.~\cite{lazaric2008transfer} transferred the state transition samples from the source to the target task by calculating the compliance (similarity) between tasks using the Bayes theorem.
Brys et al.~\cite{brys2015policy} provided an inter-task mapping based on state spaces to measure the similarity between two tasks, and transferred the value functions using reward shaping.
Song et al.~\cite{song2016measuring} transferred the value functions by measuring the distance between the source and the target tasks based on the expected models of the MDPs.
Laroche and Barlier~\cite{laroche2017transfer} estimated the reward function of the target task by reusing the experience instances of a source task.
Mustafa et al.~\cite{mustafa2021assured} presented an assured metacognitive RL-based autonomous control framework to learn to choose reward functions that satisfy desired specifications and achieve significant performance across a variety of circumstances.
However, transfer methods based on value functions or samples usually rely on well-estimated models of the MDPs or some prior knowledge of the target task for similarity measurement, which leads to high computational complexity and could be infeasible to transfer in practice.

Instead, the other kind of approach attempts to directly transfer policies learned in source tasks, which eliminates the requirement on the prior knowledge of the target task or the assumed models of the MDPs.
Fern{\'a}ndez et al.~\cite{fernandez2010probabilistic} proposed probabilistic policy reuse that transfers source policies to bias the agent's exploration strategy in the target task.
Li and Zhang~\cite{li2018optimal} developed an online method of optimally selecting source policies by formulating it as a multi-armed bandit (MAB) problem. 
Li et al.~\cite{li2018context} proposed a context-aware multi-policy reuse approach by employing the option framework to select the most appropriate source policy according to some contexts (e.g., a subset of states). 
Yang et al.~\cite{yang2020efficient} formulated the multi-policy transfer as an option learning problem, which serves as a complementary optimization objective of policy learning in the target task.
These policy reuse approaches focus on facilitating learning the optimal policy for the target task, other than a quick response to an unknown task or rapid convergence of the target policy.
Nevertheless, it is often required for RL to act online and respond quickly, in terms of rapid convergence, to novel tasks in real-world scenarios such as applications involving interactions with humans.

Rosman et al.~\cite{rosman2016bayesian} contributed BPR to quickly select source policies from an offline library using the Bayesian inference based on some observation signals and a trained observation model.
BPR prefers to quickly select a pre-learned policy from a fixed library, which does not guarantee that the appropriate policy will be learned.
Especially when the agent is faced with new unknown tasks, it may even result in negative transfer~\cite{pan2009survey}.
Further, BPR is successfully applied to handle non-stationary opponents in multi-agent systems.
Hernandez-Leal et al.~\cite{hernandez2016bayesian} proposed an extension, BPR+, to enable online learning of new models when the learning agent detects that the current policies are not performing optimally.
Zheng et al.~\cite{zheng18deep,zheng2021efficient} proposed deep BPR+ by extending BPR+ with a neural network.
A rectified belief model using the neural network approximator is introduced to achieve accurate policy detection, and a distilled policy network is proposed as the policy library to store and reuse policies efficiently.
Gao et al.~\cite{gao2022bayesian} investigated how to play with unknown opponents in bilateral negotiation games based on deep BPR+.

In contrast to the family of BPR algorithms, we use more informative observation signals such as the state transition sample, and we propose a scalable observation model to efficiently update the task belief.
Although BPR+ algorithms can incorporate new models online, they still need to retrain the observation model in an offline manner when adding a new source policy into the library.
Instead, our scalable observation model allows us to extend our method to continual learning settings conveniently in a truly online manner.

\section{Our Method}\label{method}
In this section, we propose an improved BPR method that aims at more efficient policy transfer in DRL.
First, we formulate the scalable observation model by fitting the state transition function from limited samples.
Next, we describe in detail how to use GP and NN to estimate the observation model.
Then, we extend BPR from the conventional offline mode to the continual learning setting, which can be established naturally using the scalable observation model.
Finally, we present the detailed algorithm.

\subsection{Scalable Observation Model}
In general, BPR maintains an observation model from some observation signals to update a task similarity measure, i.e., a belief, over source tasks for positive policy reuse.
Naturally, the choice of the observation signal is crucial, since it determines the granularity of the policy selection frequency and the effectiveness of the policy reuse. 
The most widely used observation signal in BPR is the episodic return $U$ when fully executing an associated policy.
However, using the episodic return as the observation signal has two weaknesses.
One is that we must wait until the end of a full episode to obtain the signal.
Some applications have very long episodes, so delaying observing the signal until the end of the episode is too slow.
The other is that the episodic return is only a scalar that contains limited information.
For example, two policies with large differences may receive similar episodic returns in the same task, or one policy can also obtain similar episodic returns in two tasks with large differences.
Instead, we use the state transition sample $(s,a,r,s')$ as the observation signal, which is supposed to reveal more task information since the state transition function $\mathcal{P}(s',r|s,a)$ completely characterizes the task's dynamics.

Suppose that we have $n$ source tasks, and we train one source policy in each task.
The observation signal could be accrued by the agent by storing the history of all state transition samples encountered during the execution of all $n$ source policies in all $n$ source tasks.
The observation model $\mathrm{P}(\sigma|\tau, \pi)$, in this case, is a tabular-based empirical estimate of the expected state transition function of the MDPs, which could be expensive and even infeasible to learn and maintain.
It requires a large amount of state transition samples to estimate the probability distribution of the observation model, and will encounter the ``curse of dimensionality" in large or continuous state-action spaces.
Additionally, this may not generalize well, especially in cases with sparse sampling.
When encountering a new sample that has not been recorded by the observation model, the Bayesian update of the task belief can easily be inaccurate in the policy reuse phase.
Since the performance of BPR highly depends on the inference of the task belief, a scalable approach is necessary for building the observation model using limited state transition samples as the signal.

To address the above challenges, we propose a scalable observation model based on fitting the state transition functions of sources tasks $\mathcal{P}_j(s',r|s,a)$ from a small number of state transition samples $\mathcal{D}_j=\{(s_i^j,a_i^j,r_i^j,{s'}_i^j)\}_{i=1}^{N_j}$, where $N_j$ is the number of samples in the $j$-th source task.
For the convenience of expression, we denote the input and output of this supervised learning instance of fitting the state transition function as $x=(s,a)$ and $y=(r,s')$. 
Assume that we have learned the state transition functions of the $n$ source tasks as $\{\mathcal{P}_j(y^j|x^j)\}_{j=1}^n$.
Then, in the target task, we obtain the observation signal, i.e., a couple of state transition samples $\sigma^t=\{(x_i^0,y_i^0)\}_{i=1}^{N_0}$, by applying a selected policy $\pi_t$ on the target task, where $N_0$ is the number of samples in the target task.
The observation model $\mathrm{P}(\sigma|\tau, \pi)$ can be naturally constructed using the fitted task dynamics as 
\begin{equation}\label{obs_model}
    \mathrm{P}(\sigma^t|\tau_j, \pi_t) = \Pi_{i=1}^{N_0}\mathcal{P}_j(y_i^0|x_i^0),~~~j=1,...,n.
\end{equation}
Intuitively, $\mathcal{P}_j(y_i^0|x_i^0)$ indicates the likelihood of the learned task dynamics $\mathcal{P}_j$ fitting the sample $(x_i^0,y_i^0)$, or the degree of samples in the target task $\tau_0$ resembling those in the source task $\tau_j$.
We only need to store a small number of samples to learn the state transition functions of source tasks, which are utilized to compute the scalable observation model that can generalize to any signals observed in the target task.

\subsection{Estimation of Observation Model}
Since $\{\mathcal{P}_j\}_{j=1}^n$ are unknown,  we only have access to an estimation of the observation model in~(\ref{obs_model}).
To obtain an approximation of the unknown densities, we employ two typical supervised learning approaches, the non-parametric GP~\cite{seeger2004gaussian} and the parametric NN~\cite{lecun2015deep}, to fit the state transition function for each source task.
These two methods are suitable for different situations with respective advantages.

First, we adopt GP due to its sample efficiency and the ability to provide uncertainty measurements on the predictions.
GP is a non-parametric and Bayesian approach to regression that has been successfully adopted in many existing works in the RL community~\cite{rasmussen2003gaussian,deisenroth2011pilco,doshi2016hidden,berkenkamp2018safe}. For example, PILCO~\cite{deisenroth2011pilco} used the GP to approximate the state transition function, which was employed to assist the long-term planning and policy evaluation in the context of model-based RL.

Given a test point $x_*$, the $j$-th GP returns a Gaussian distribution over the output's mean, i.e., $\hat{y}_*\sim\mathcal{N}(\mu_{GP_j}(x_*), \operatorname{cov}_{GP_j}(x_*))$, as
\begin{equation}
\label{mu}
\begin{aligned}
\mu_{GP_j}\left({x}_{*}\right) &={\kappa}_{*j}\left(\kappa_{jj} +\varepsilon^2  {I} \right)^{-1} {Y_j}, \\
\operatorname{cov}_{GP_j}\left({x}_{*}\right) &=\kappa_{**}-{\kappa}_{*j}\left({\kappa}_{jj} +\varepsilon^2 {I} \right)^{-1} {\kappa}_{j*},
\end{aligned}
\end{equation}
where $(X_j,Y_j)$ are the training samples $\{(x_i^j,y_i^j)\}_{i=1}^{N_j}$ from the $j$-th source task, and $\kappa$ is the kernel function such that ${\kappa}_{jj}$ = $\kappa(X_j, X_j)$, ${\kappa}_{*j}=\kappa (x_*,X_j)$, ${\kappa}_{j*}=\kappa (X_j,x_*)$, ${\kappa}_{**}=\kappa (x_*,x_*)$.
The most commonly used kernel function is the gaussian kernel, also known as the radial basis function (RBF):
\begin{equation}
\kappa\left(x, x^{\prime}\right)=\delta^{2} \exp \left(-\frac{\left\|x-x^{\prime}\right\|_{2}^{2}}{2 l^{2}}\right),
\end{equation}
where $\delta$ and $l$ are the hyper-parameters of the RBF kernel, representing its signal variance and characteristic length scale.
For more commonly used kernels, please refer to~\cite{seeger2004gaussian}.
Considering each sample as an independent Gaussian distribution, the observation model in~(\ref{obs_model}) can be re-arranged using the $n$ fitted GPs $\{GP_j\}_{j=1}^n$ as
\begin{equation}\label{gp_obs}
    \mathrm{P}\left(\sigma^t|\tau_j, \pi_t\right) = \Pi_{i=1}^{N_0}\mathcal{N}\left(y_i^0; \mu_{j}(x_i^0), \operatorname{cov}_{j}(x_i^0)+\varepsilon_{GP}^2 \right),
\end{equation}
where $\mu_j$ and $\operatorname{cov}_j$ denote $\mu_{GP_j}$ and $\operatorname{cov}_{GP_j}$ for simplicity, and $\varepsilon_{GP}^2$ is a constant additional variance of the Gaussian distribution, which is used to enhance generalization of GP.

Second, we adopt NN as another technique to fit the state transition function since it can scale to extremely high-dimensional problems.
Having been widely investigated in supervised learning tasks, NN can extract useful information from very large data sets to build extremely complex models.
Naturally, we parameterize the state transition function of a given source task as 
\begin{equation}\label{s}
\hat{Y}_j=g_{\vartheta_{j}}(X_j),
\end{equation}
where $g_{\vartheta_j}(\cdot)$ is a neural network parameterized by weights $\vartheta_j$, and $\hat{Y}_j$ is the predicted output of the network.
Each neural network is trained in a supervised learning way to minimize the loss function, e.g., the mean squared error (MSE), as
\begin{equation}
    \mathcal{L}(\vartheta_j) = (Y_j - \hat{Y}_j)^2.
\end{equation}
With the $n$ fitted NNs, the observation model in~(\ref{obs_model}) represents each sample as an independent Gaussian distribution, such that: 
\begin{equation}\label{nn_obs}
    \mathrm{P}\left(\sigma^t|\tau_j, \pi_t\right) = \Pi_{i=1}^{N_0}\mathcal{N}\left(y_i^0; g_{\vartheta_{j}}(x_i^0),  \varepsilon_{NN}^2 \right),
\end{equation}
where the preset constant $\varepsilon_{NN}^2$ is the variance of the Gaussian distribution.
 
For the sake of simplicity, let $f_{j}(x_i^0)$ and $\xi_{j}^2$ denote the mean and variance of the Gaussian distribution in the observation model, respectively.
Then, we can obtain a uniform formation for the observation models in (\ref{gp_obs}) and (\ref{nn_obs}) as
\begin{equation}\label{uniform_obs}
    \mathrm{P}\left(\sigma^t|\tau_j, \pi_t\right) = \Pi_{i=1}^{N_0}\mathcal{N}\left(y_i^0; f_{j}(x_i^0),  \xi_{j}^2 \right),~j=1,...,n.
\end{equation}
With the estimated observation model, the task belief $\beta^t$ in (\ref{bpr}) can be efficiently updated using the Bayes' rule as
\begin{equation}\label{uniform_belief}
    \beta^t(\tau_j)=\frac{\prod\limits_{i=1}^{N_0}\mathcal{N}\left(y_i^0;  f_{j}(x_i^0), \xi_{j}^2 \right)\beta^{t-1}(\tau_j)}{\sum\limits_{j'=1}^{n} \prod\limits_{i=1}^{N_0}\mathcal{N}\left(y_i^0; f_{j'}(x_i^0), \xi_{j'}^2 \right)\beta^{t-1}(\tau_{j'})}.
\end{equation}

\subsection{Extension to Continual Learning} 
The conventional BPR algorithm is performed in an \textit{offline} manner.
It has to acquire a collection of pre-learned behaviors in advance, and also needs to apply a fixed set of policies on all source tasks to obtain a tabular-based observation model offline.
Nevertheless, when encountering a new task, it is likely that none of the policies in the library is suitable, which leads to negative transfer.
Although the BPR+ algorithms can incorporate new models online, when adding a new source policy into the library, they still need to re-estimate the observation model by applying all source policies on the new task and applying the new policy on all tasks.
This can be a big barrier to the practicality of the algorithm due to the expensive and inefficient update of the observation model.
Moreover, different from the learned policies, the source tasks may be inaccessible in the new learning process of the target task, in which case the observation model is infeasible to update.


To achieve artificial general intelligence, RL agents should constantly build more complex skills and scaffold their knowledge about the world in a \textit{continual learning} manner.
In the paper, we extend our method to the continual learning setting such that in the policy reuse phase, the learning agent can consult the stored library first, and either retrieve the most suitable policy from the library or expand a new policy into the library.
When incorporating a new policy, we only need another GP or NN to fit the state transition function of the corresponding new source task $\mathcal{P}_{n+1}$ using a small number of samples from that task, independent of the policies and fitted GPs or NNs from the existing source tasks.
The proposed observation model in~(\ref{uniform_obs}) is naturally scalable to the continual expansion of a new source task in a plug-and-play fashion as
\begin{equation}
    \mathrm{P}\left(\sigma^t|\tau_{n+1}, \pi_t\right) = \Pi_{i=1}^{N_0}\mathcal{N}\left(y_i^0; f_{n+1}(x_i^0), \xi_{n+1}^2 \right).
\end{equation}

Note that the conventional BPR also maintains a performance model, i.e., a distribution of returns from each policy on the source tasks, which is utilized together with the task belief to select the most appropriate policy for reusing.
As we observe the state transition function that completely characterizes the task's dynamics, the scalable observation model is informative to capture the task similarity effectively.
We can merely rely on the task belief in~(\ref{uniform_belief}) to choose reuse policies, and hence we abandon the use of the performance model for better applicability in the continual learning setting.

\subsection{Algorithm}
Based on the above statements, we present the general form of our method in Algorithm~\ref{algo}. 
It mainly consists of two phases: the reuse phase that selects the most appropriate policy from the library in Lines~2-13, and the learning phase which expands a new optimal policy into the library in Lines~14-17.

\begin{algorithm}[tb]
\caption{Efficient BPR with Scalable Observation Model}
\label{algo}
\KwIn{
      Source task set $\mathcal{T}$, \newline 
      source policy library $\Pi$ and GPs/NNs, \newline 
      number of episodes $K$, \newline
      the maximum number of steps $M$.}
Initialize belief $\beta^{0}(\mathcal{T})$ with a uniform distribution \\
\uIf {execute a reuse phase}{
    \While{episode $\leq$ $K$}{
		\While{step $\leq$ $M$  $\&$ not reaching the goal}{
		    Select  a policy $\pi^{t} \in \Pi$ based on $\beta^{t-1}$\\
		    Apply $\pi^{t}$ on target task $\tau_{0}$ and receive observation signal $\sigma^{t}$ \\
		    Estimate $\mathrm{P}(\sigma^{t}|\mathcal{T}, \pi^t)$ in~(\ref{uniform_obs}) by applying $\sigma^{t}$ on source GPs or NNs \\
		    Update task belief $\beta^t$ in~(\ref{uniform_belief}) using the estimated $\mathrm{P}(\sigma^{t}|\mathcal{T}, \pi^t)$
		    }
   
		 \If {a sufficiently different task is detected}{Switch to the learning phase }  

	}}

\ElseIf {execute a learning phase} {
			 Learn a new policy $\pi_{0}$ and fit a new GP/NN for $\tau_0$ \\
			 Expand $\pi_0$ as $\pi_{n+1}$ into the library \\
			 Switch to the reuse phase 
}

\end{algorithm}

First, the policy reuse phase is performed when facing the target task.
At update step $t$ (the step size is $N_0$), the agent selects a reuse policy $\pi^t$ from the library $\{\pi_j\}_{j=1}^n$ according to the task belief $\beta^{t-1}$, and receives the observation signal $\sigma^{t}$ after applying the selected policy $\pi^t$ on the target task $\tau_{0}$.
Next, the scalable observation model $\mathrm{P}(\sigma^t|\tau_j, \pi^t)$ in~(\ref{uniform_obs}) can be efficiently estimated by applying the obtained signal $\sigma^{t}$ on the fitted GPs or NNs from the source tasks, which is used to update the task belief in~(\ref{uniform_belief}).
The two processes are iteratively alternated within a learning episode until the goal or the maximum number of steps is reached.
At the end of the episode, we check whether the target task is sufficiently different from all source tasks according to the average return $\bar{U}$ over $k$ episodes.
If $\bar{U}$ is below a preset threshold, the agent will switch to the learning phase in subsequent episodes where a new optimal policy $\pi_{n+1}$ is learned for the target task using any DRL algorithm.
The obtained new policy $\pi_{n+1}$ is expanded into the policy library, together with the target task $\tau_0$ being extended as a new source task $\tau_{n+1}$.
Accordingly, a new GP or NN is generated to fit the dynamics of the new source task using a small number of samples collected from the learning phase. Specifically, for GP, we store a small number of samples for the new source task to calculate the probability distribution of the observation signals when fitting the new GP. For NN, we need to use the collected samples to train a new NN model to fit the state transition function for the new source task.
Finally, the agent switches back to the reuse phase for the next policy transfer problem.

Note that in some practical applications, we can empirically choose $(s,a,r)$ or $(s,a,s')$ alone as the observation signal for ease of use, in which case the output $y$ of the task dynamics function in~(\ref{obs_model}) is the reward $r$ or the next state $s'$.
Furthermore, $N_{0}$ determines the granularity of the policy selection frequency, in this paper, we update the policy reusing strategy every time step for better sample efficiency, i.e., $N_0=1$.
In summary, our algorithm estimates the observation model from limited samples to realize efficient policy transfer.
Meantime, using the scalable observation model, our algorithm can be easily extended to the continual learning setting, which makes it more practical in real-world scenarios.

\section{Experiments}\label{experiments}
In this section, we evaluate our method on five different domains with increasing levels of complexity. 
Through these experiments, we aim to build problem settings that are representative of the types of transfer learning that RL agents may encounter in real-world scenarios.
Our experiments are mainly divided into three parts:
\begin{enumerate}
    \item First, we implement our GP-based method in three relatively simple domains to demonstrate that it can achieve efficient policy transfer when faced with target tasks that are similar to source tasks.
    The results are shown in Section~\ref{GP-experiment}.
    \item 
    Second, we implement our NN-based method in two high-dimensional complex domains from MuJoCo, to verify that our method enables efficient policy transfer for more sophisticated tasks.
    The results are shown in Section~\ref{NN-experiment}.
    \item Finally, we apply our method to all domains to prove that compared to the basic BPR algorithm, it can achieve continual learning and thus avoid negative transfer when faced with target tasks that are very different from the source tasks, as shown in Section~\ref{Continual-experiment}.
\end{enumerate}

In Sections~\ref{GP-experiment} and~\ref{NN-experiment}, we compare our method to three baseline approaches for transferring policies: the family of BPR algorithms using episodic return as the observation signal~\cite{rosman2016bayesian,zheng2021efficient}, the probabilistic Policy Reuse with DRL (PR-DRL) algorithm~\cite{fernandez2010probabilistic,barreto2016successor,wang2019tnnls}, and the Optimal Policy Selection with DRL (OPS-DRL) algorithm~\cite{li2018optimal}.
In Section~\ref{Continual-experiment}, we evaluate our method in comparison to the basic BPR  algorithm~\cite{rosman2016bayesian}, and the settings of the BPR algorithm are consistent with those in Sections~\ref{GP-experiment} and~\ref{NN-experiment}.
For a fair comparison, we make some improvements to the baselines so that they can be directly compared with our method.
The details of the three baselines are given in the Appendix.

All experimental results are averaged over 10 trials. 
The shaded area represents the 95\% confidence interval for evaluation curves, and the standard errors are presented for numerical results.
All the algorithms are implemented with Python 3.5 running on Ubuntu 16 with 48 Intel Xeon E5-2650 2.20GHz CPU processors, 193-GB RAM, and an NVIDIA Tesla GPU of 32-GB memory.

\subsection{Results for Efficient Policy Transfer Based on GP}\label{GP-experiment}
\begin{figure*}[tb]\centering
 \subfigure[]{\includegraphics[width=0.45\columnwidth]{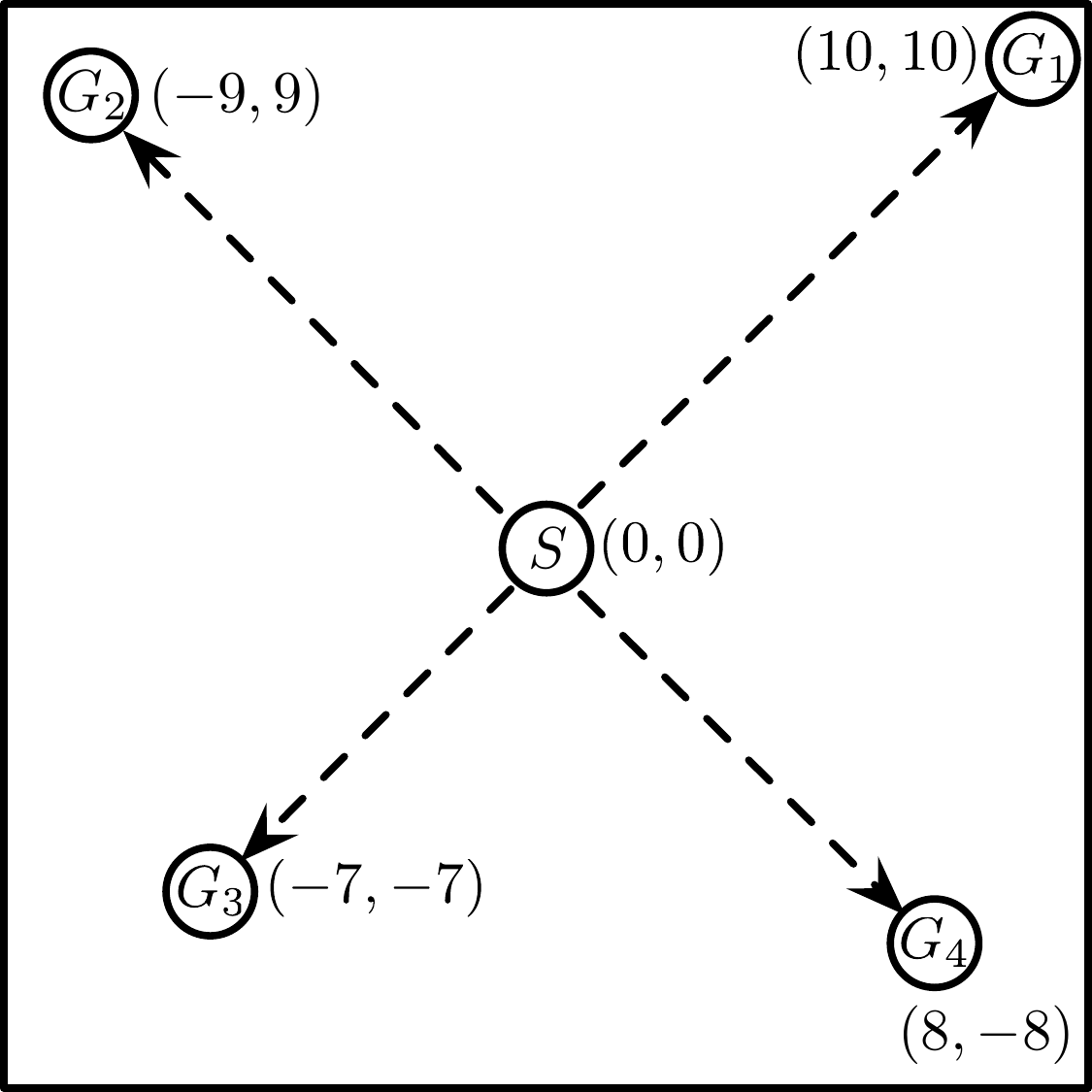}}\hspace{1em}
 \subfigure[]{\includegraphics[width=0.45\columnwidth]{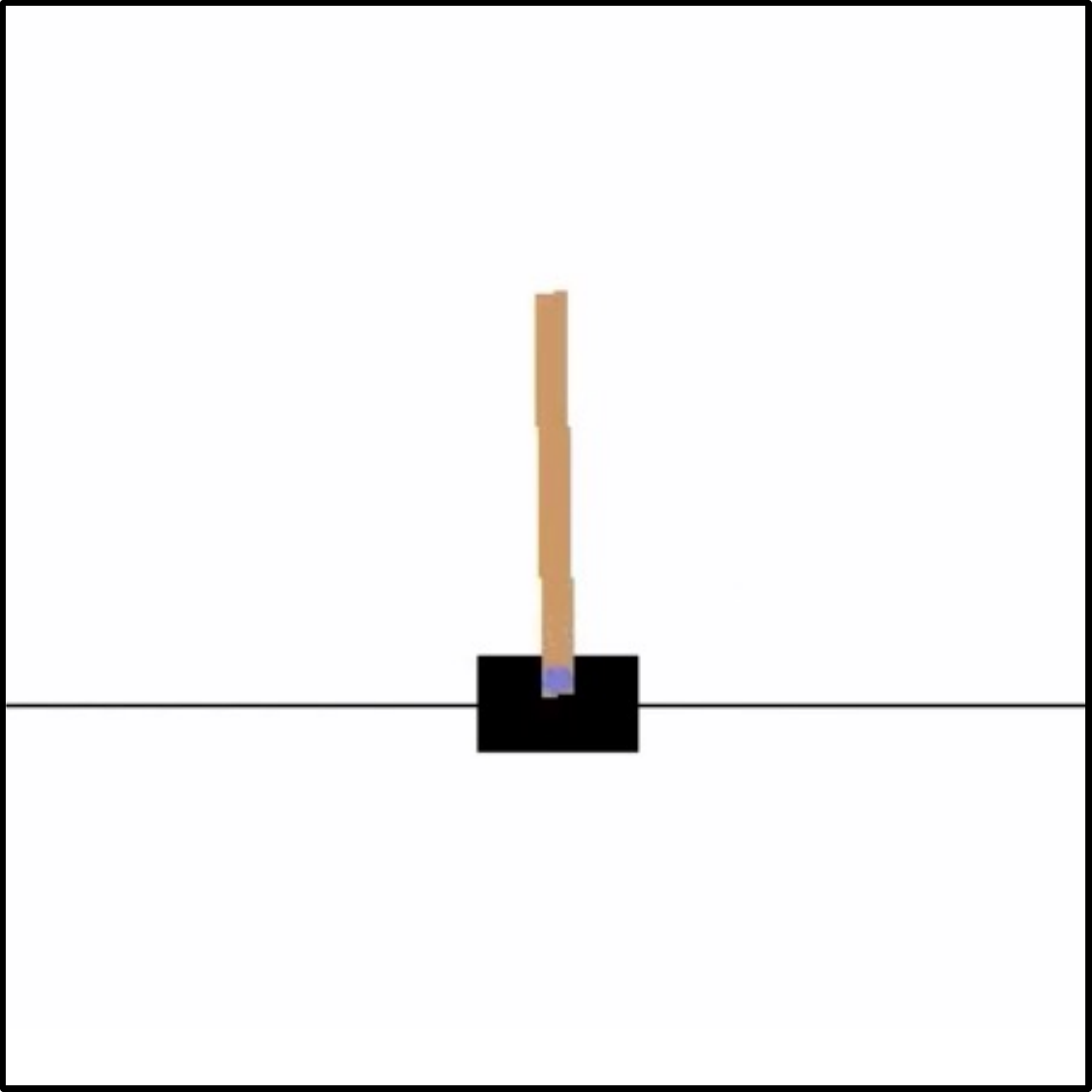}} \hspace{1em}
 \subfigure[]{\includegraphics[width=0.45\columnwidth]{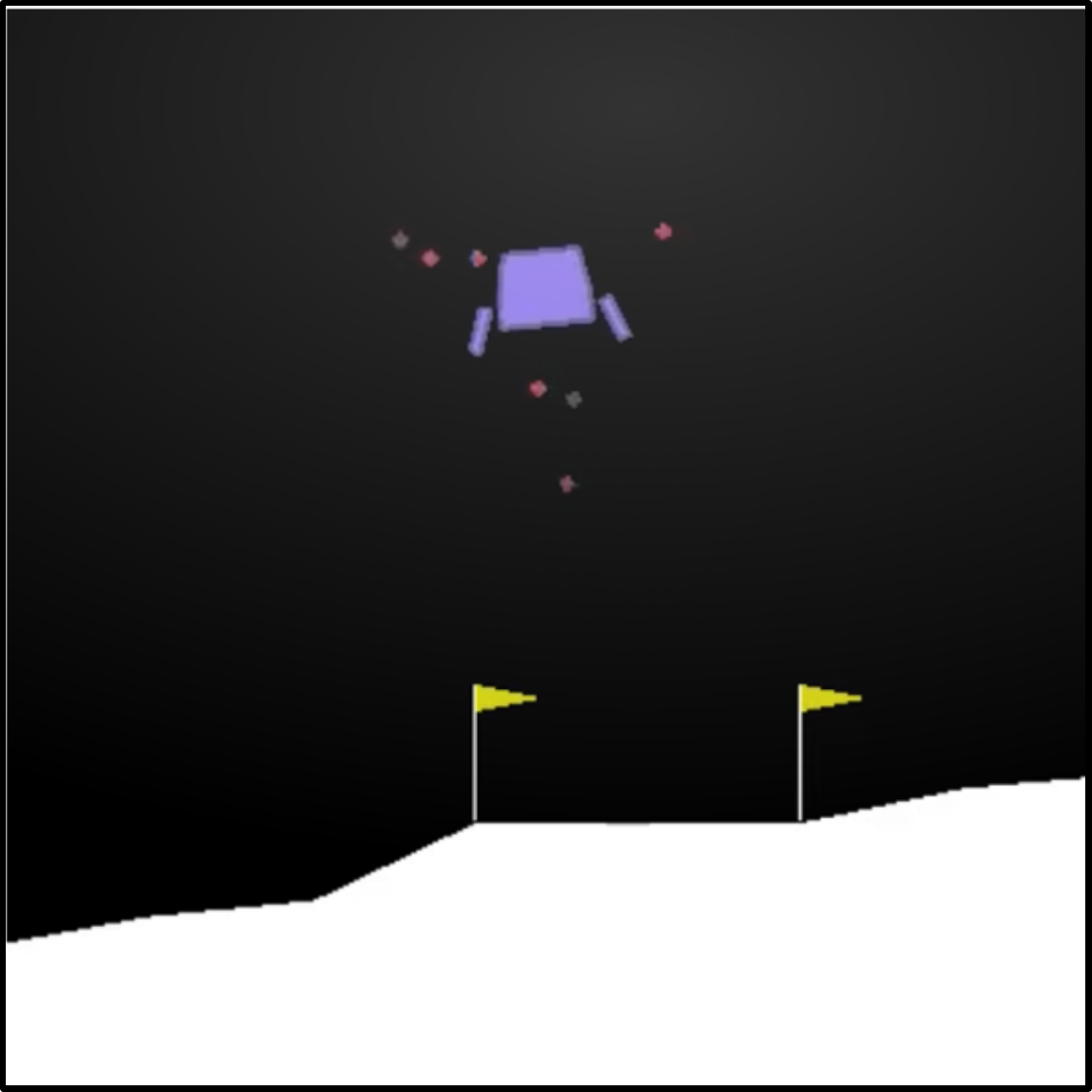}}
 \caption{Three types of dynamic environments. (a) 2-D Navigation domain, S is the start point and G is the goal point. (b) CartPole domain. (c) LunarLander domain.}
 \label{fig1-env} 
\end{figure*}

In the experiment of this section, we choose three representative types of dynamic environments, as shown in Fig.~\ref{fig1-env}, and the details are as follows.

\subsubsection{2-D Navigation}
We first consider the navigation domain where an agent must move to a goal position in a 2-D surface, which is a continuous-state and continuous-action problem.
The states are the current 2-D positions and the actions are two-dimensional vectors clipped to be in the range of $[-1, +1]$. 
Episodes terminate when the agent is within 0.5 of the goal or reaches the maximum number of steps $M = 100$. The reward is the negative Euclidean distance to the goal minus a controlled cost that is positively related to the scale of actions. 
We consider four source tasks, where they have the same starting points $(0,0)$ and different goal positions as $(10,10)$, $(-9,9)$, $(-7,-7)$, $(8,-8)$, respectively.
In this domain, tasks only differ in reward functions.
Therefore, we employ the tuple ($s, a, r$) as the observation signal to fit the GP and estimate the scalable observation model.

\subsubsection{CartPole}
We next consider a classical control problem with a continuous state space and a discrete action space, described in \cite{geva1993cartpole} and implemented experiments in OpenAI Gym \cite{brockman2016openai}.
The states are four-dimensional vectors, and the actions are two discrete values of 0 and 1.
Episodes terminate when the pole falls or reaches the maximum number of steps $M = 100$.
In order to encourage the agent to balance the pole, a positive reward of +1 is given at every step when the angle between the pole and vertical line is smaller than a small threshold.
Otherwise, the reward is 0.
The system is controlled by applying a constant force $F=10$ newtons to the cart, and the agent can apply full force to the cart in either direction, i.e., two possible actions.
We set up two source tasks by adding a constant force $F'$ with a fixed direction on the cart.
The agent needs to balance the pole with the interference of $F'=5$ newtons in one source task, and with $F'=-5$ in the other task.
In this domain, 
the tuple $\left(s, a, s^{\prime}\right)$ is set as the observation signal to infer the most appropriate policy for the target task.

\subsubsection{LunarLander}
We choose a relatively complex task, the LunarLanderContinuous-v2 domain from OpenAI-Gym, which involves landing a spacecraft safely on a lunar surface.
This is a high-dimensional continuous control problem with sparse reward, and is representative of real-world problems where it is considerably difficult to learn accurate dynamics.
The states are eight-dimensional vectors.
The actions are two-dimensional vectors clipped to be in the range of $[-1, +1]$, which are used to control the powers of the main and side engines. 
Episodes finish if the lander crashes or comes to rest or reaches the maximum number of steps $M = 1000$.
In this case, it can take a large number of steps to reach the goal, and the reward is significantly delayed until the end of the long episode.
We set up three source tasks that represent three typical scenarios by applying an additional constant power with a fixed direction on the main engine of the spacecraft. 
The additive powers are $(0.5, 0)$, $(-0.5, 0)$, and $(0, 0.5)$, respectively.
A successful transfer agent should learn to transfer skills from the source tasks and try to avoid the potential risk of landing the spacecraft. 
Here, the moon's surface is generated randomly in each episode, so the environmental dynamics are learned on noisy data. 
To solve this problem, during the experiment, we set up a constant random seed for the environment. 
In this setting, tasks mainly differ in the state transition functions.
Therefore, we can employ the tuple $\left(s, a, s^{\prime}\right)$ as the observation signal to infer the task belief.


In this experiment, we select some target tasks that are relatively close to the source tasks in each domain, which ensures that there are suitable policies for the target tasks in the policy library.
\begin{itemize}
    \item In 2-D Navigation domain, we select twelve target tasks, and their goal positions are ($10.5, 10$), ($10, 9.5$), ($-8.5, 9$), ($-9, 9.5$), ($-6.5,-7$), ($-7,-7.5$), ($7.5,-8$), ($8, -7.5$), ($10, 10$), ($-9, 9$), ($-7,-7$), ($8,-8$).
    \item In CartPole domain, we select six target tasks, and their values of the additional force $F^{\prime}$ are set as $4.5$, $5$, $5.5$, $-5.5$, $-5$, $-4.5$.
    \item In LunarLander domain, we select nine target tasks, and their value of additive powers are ($0.45, 0$), ($-0.45, 0$), ($0, 0.45$), ($0.55, 0$), ($-0.55, 0$), ($0, 0.55$), ($0.5, 0$), ($-0.5, 0$), ($0, 0.5$).
\end{itemize}

In each experiment, the number of learning episodes in the target task, $K$, is set as 10, which is supposed to be sufficient for all tested methods to converge to the most appropriate policy with the highest return.
The hyperparameters are set as: $\delta = 1$  and $l = 2 $ for the RBF kernel of the GPR, and $\varepsilon^2_{GP} = 0.1$ for the Gaussian distribution of all test domains.
And in this experiment, we use vanilla policy gradient (REINFORCE)~\cite{williams1992simple,duan2016benchmarking} and twin delayed deep deterministic policy gradient (TD3) algorithm~\cite{fujimoto2018addressing} to learn optimal policies for source tasks.

\begin{figure*}[tb]\centering
 \subfigure[]{\includegraphics[width=0.32\textwidth]{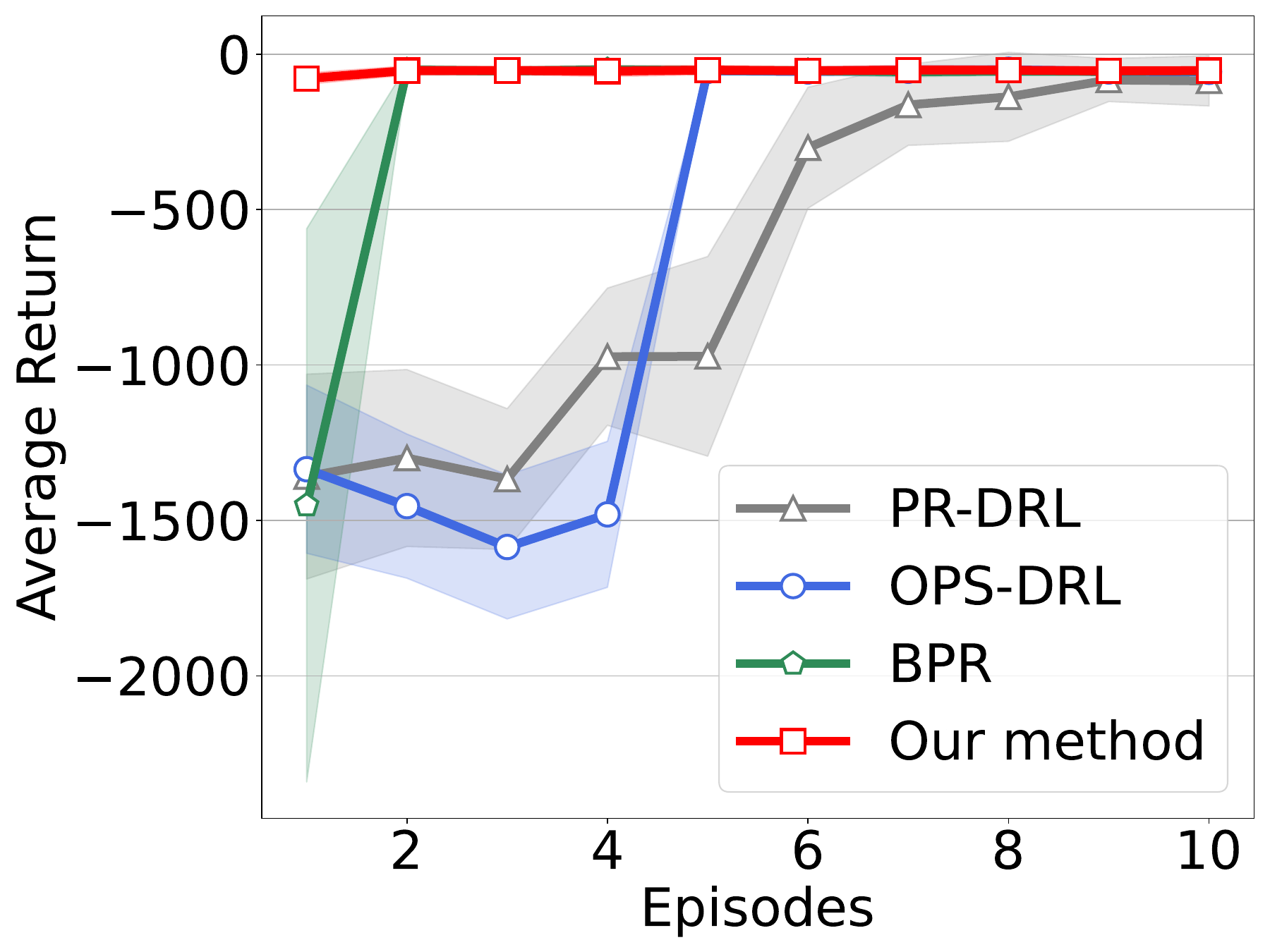}}
 \subfigure[]{\includegraphics[width=0.32\textwidth]{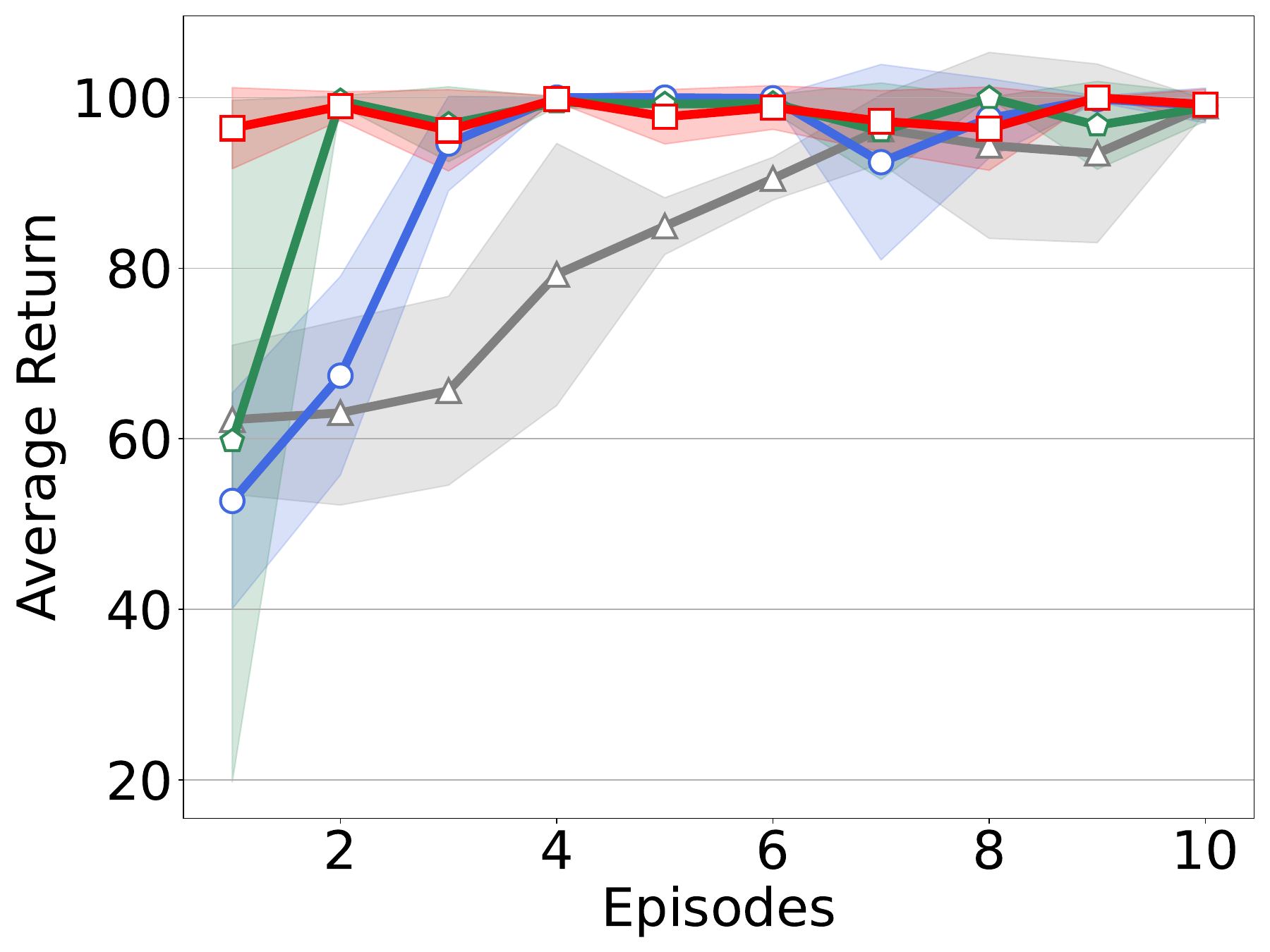}}
 \subfigure[]{\includegraphics[width=0.32\textwidth]{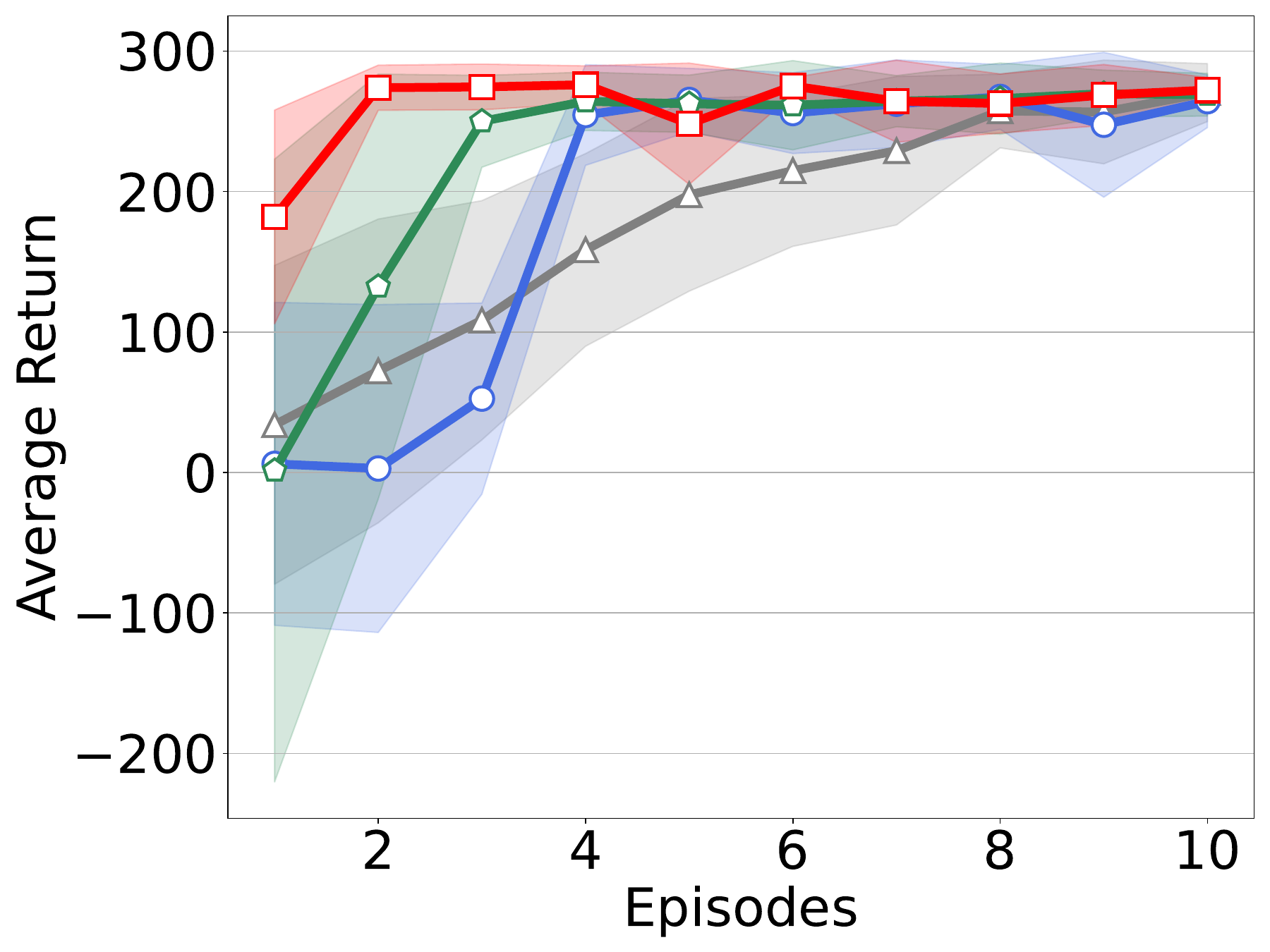}}
 \caption{Average return per episode of all tested methods in: (a) 2-D Navigation; (b) CartPole; (c) LunarLander. }
 \label{fig1}
\end{figure*}

We present the primary results of our GP-based method and all baseline approaches on three relatively simple experimental domains, 
and all the results are averaged over multiple target tasks.
Fig.~\ref{fig1} shows the average return per learning episode, and Table~\ref{tab1} reports the numerical results in terms of average return over all learning episodes.
BPR obtains better policy transfer performance than PR-DRL and OPS-DRL, which is supposed to benefit from leveraging the efficiency of the Bayesian inference framework.
It is clear that our method achieves much more efficient policy transfer in all tasks compared to baselines. 
From Figs.~\ref{fig1}-(a) and~\ref{fig1}-(b), it is observed that the received return of our method in the first episode is nearly equal to that of the best policies for target tasks, since our method can converge to the most appropriate policy using a few steps within the first episode.
In more complex domains, our method can also achieve much better jump-start performance compared to all baselines, as illustrated in Fig.~\ref{fig1}-(c). 
Further, from the numerical results in Table~\ref{tab1}, our method generally obtains the largest average return over all learning episodes in all domains.
In addition, it can be observed from the statistical results that our method mostly obtains smaller confidence intervals and standard errors than the baselines.
This phenomenon indicates that our method can provide more stable source task selection and knowledge transfer.

\begin{table*}[tb]
\centering
\caption{Average return over all episodes of all test methods implemented in three domains.}
\renewcommand\arraystretch{1.2}
\begin{tabular}{c|c|c|c|c}
\cmidrule[\heavyrulewidth]{1-5}
Method & PR-DRL & OPS-DRL &  BPR & Our method\\
\hline

2-D Navigation         & $-674.19\pm 581.44$               & $-618.02 \pm 709.72$ &  $-193.39\pm 505.16$    & $\mathbf{-55.33 \pm 16.53}$ \\

CartPole   & $82.88\pm 16.41$       & $90.36\pm 17.16$ &  $94.59 \pm 17.46$    & $\mathbf{98.08\pm 3.53}$ \\

LunarLander       &$179.88\pm  105.79$      &  $187.87 \pm 126.69$ &  $224.09\pm 121.28$    & $\mathbf{259.75\pm 41.98}$ \\

\cmidrule[\heavyrulewidth]{1-5}
\end{tabular}
\label{tab1}
\end{table*}

\begin{figure*}[tb]\centering
 \subfigure[]{\includegraphics[width=0.32\textwidth]{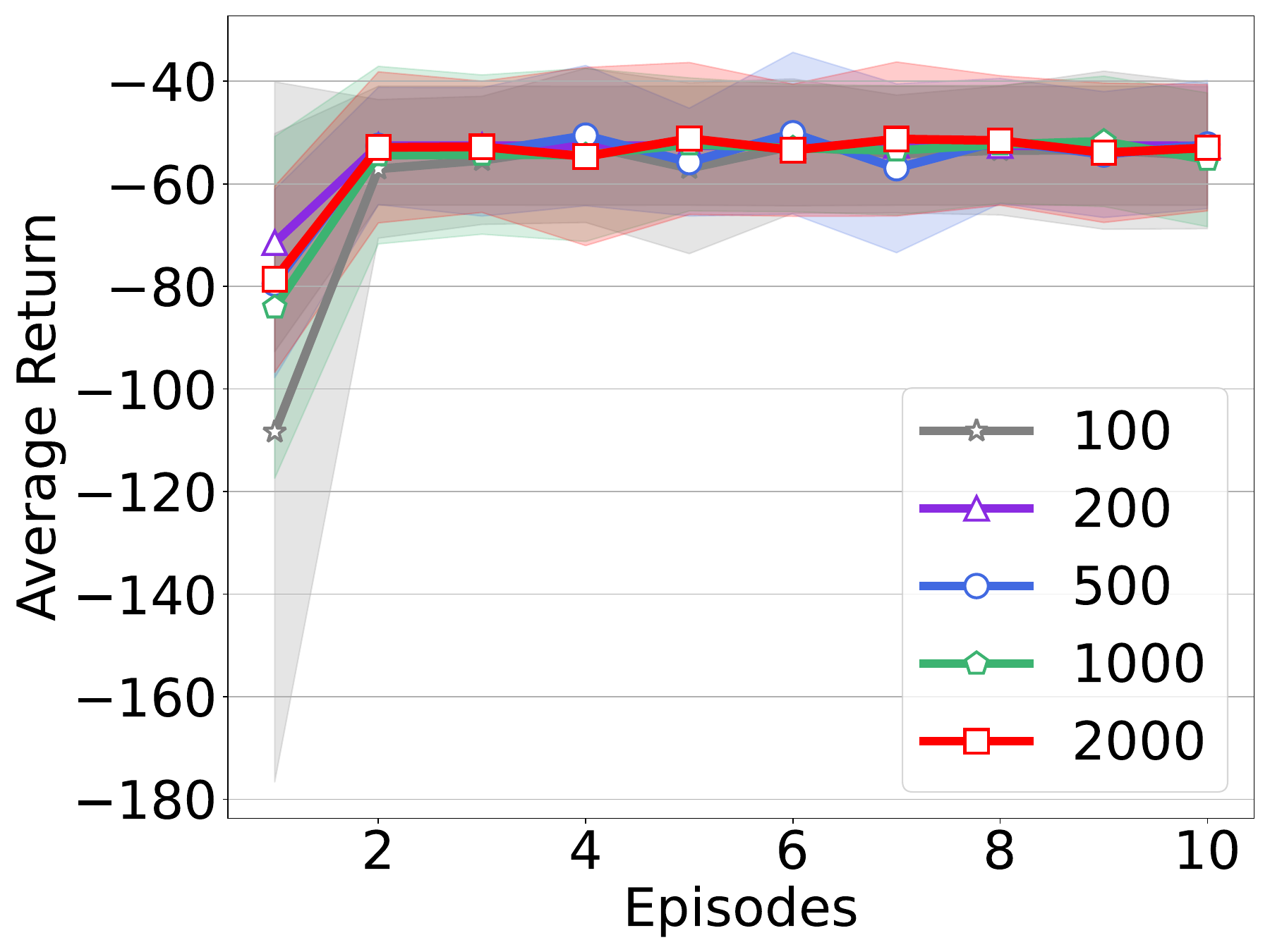}}
 \subfigure[]{\includegraphics[width=0.32\textwidth]{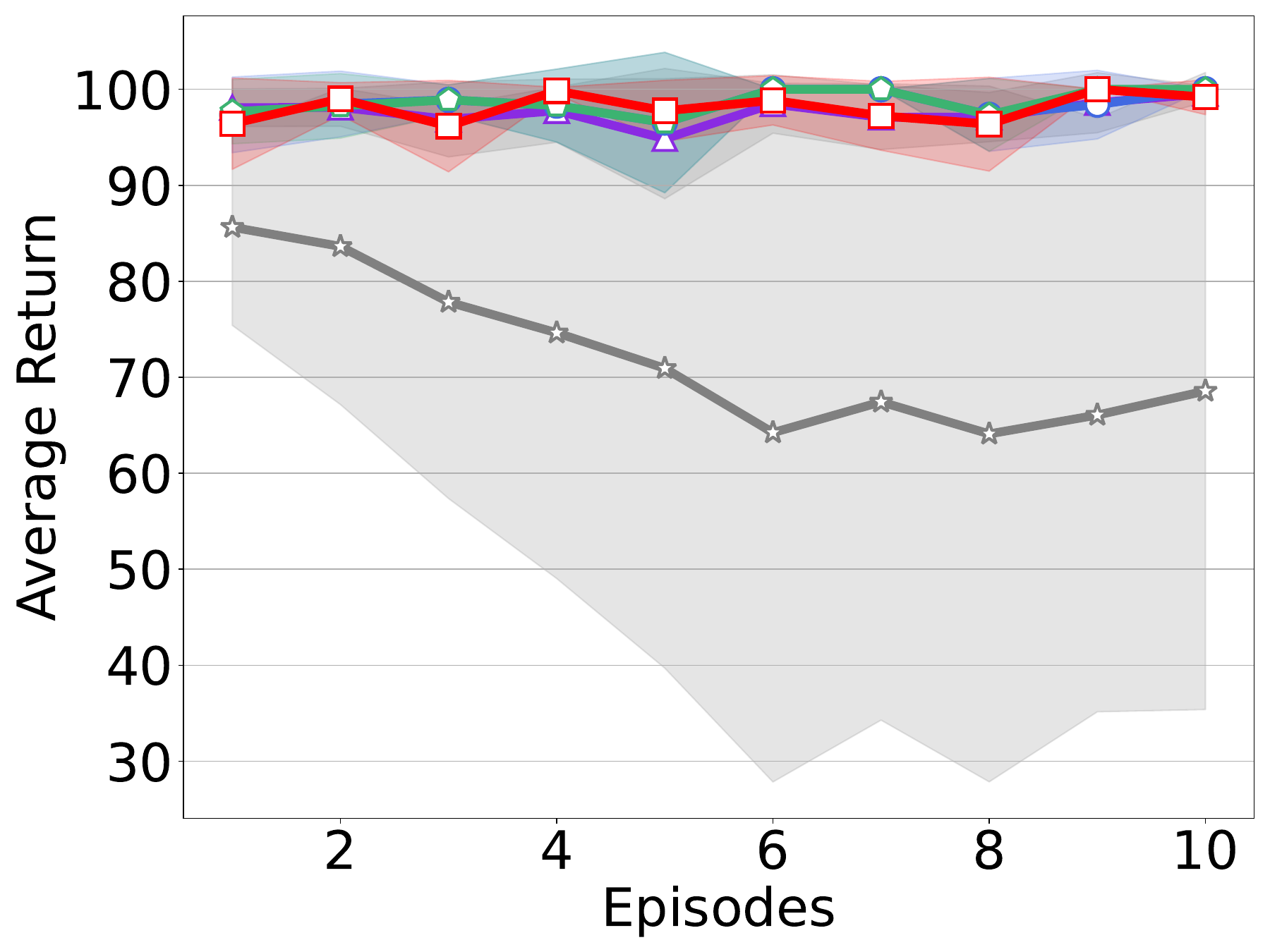}}
 \subfigure[]{\includegraphics[width=0.32\textwidth]{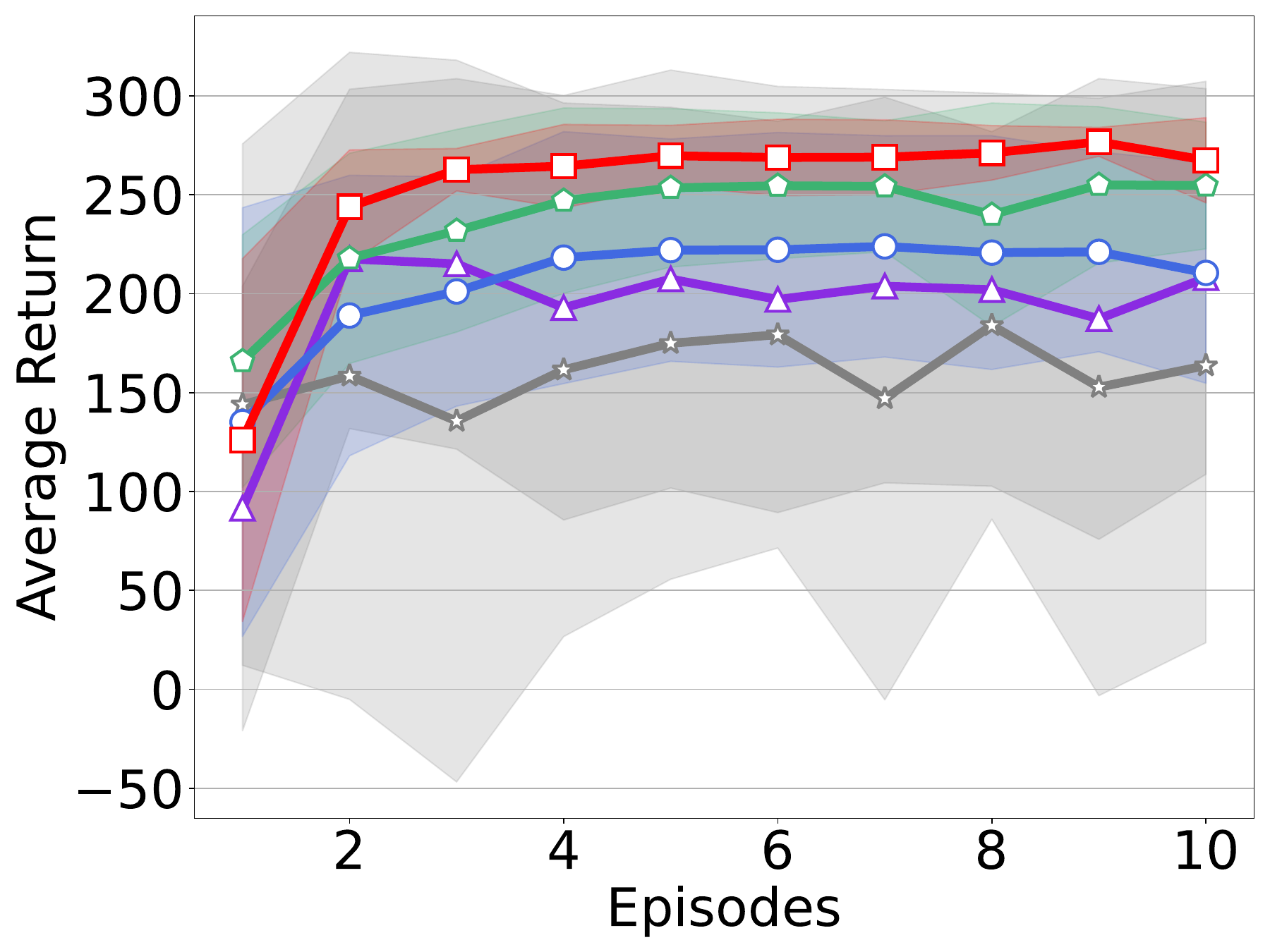}}
 \caption{Average return per episode of different sample sizes used for GPs in: (a) 2-D Navigation; (b) CartPole; (c) LunarLander.}
 \label{fig2}
\end{figure*}

\begin{table*}[tb]
\caption{Average return over all episodes of different sample sizes used for GPs in three domains.}
\centering
\renewcommand\arraystretch{1.2}
\begin{tabular}{c|c|c|c|c|c}
\cmidrule[\heavyrulewidth]{1-6}
Sample size & 100 & 200 &  500 & 1000 & 2000\\
\hline
 2-D Navigation   & $-59.87\pm 30.06$       & $\mathbf{-54.46 \pm 14.09}$ &  $-55.75 \pm 16.01$    & $ -56.53\pm 19.33$ & ${-55.33\pm 16.53}$ \\

 CartPole      & $72.31\pm 29.62$       & $97.65\pm 3.54$ &  $98.53 \pm 3.72$    & $ \mathbf{98.70}\pm 3.50$ & ${98.08\pm 3.53}$ \\
 
 LunarLander         & $186.43\pm 105.72$       & $192.43\pm 125.87$ &  $206.37\pm 70.66$     &  $237.42\pm 53.50$  & $\mathbf{251.97\pm 54.57}$ \\

\cmidrule[\heavyrulewidth]{1-6}
\end{tabular}
\label{tab2}
\end{table*}

We also study how the number of state transition samples $(s,a,r,s')$ used for fitting the GPs affects the performance of our method, i.e., identifying the relationship between the sample size and the accuracy of policy detection during the policy reuse phase in our method. 
We use $100$, $200$, $500$, $1000$, and $2000$ samples, respectively, for fitting state transition functions using GPs in these domains to observe the performance of our method.
Fig.~\ref{fig2} shows the average return per learning episode, and Table~\ref{tab2} reports the numerical results in terms of average return over all learning episodes.
In all domains, our method obtains a significant performance improvement when the sample size goes from 100 to 200. 
Then, as the sample size continues to grow, the performance of our method is slightly improved only, and the degree of performance improvement is decreasing. 
Especially in the 2-D Navigation and CartPole domains, the performance of our method remains roughly the same.
It demonstrates that by using only a small number of samples to fit the state transition function, our method is capable of accurately estimating the scalable observation model and inferring the task belief for efficient policy transfer.

\subsection{Results for Efficient Policy Transfer Based on NN}\label{NN-experiment}
\begin{figure}[tb]\centering
\subfigure[]{\includegraphics[width=0.4\columnwidth]{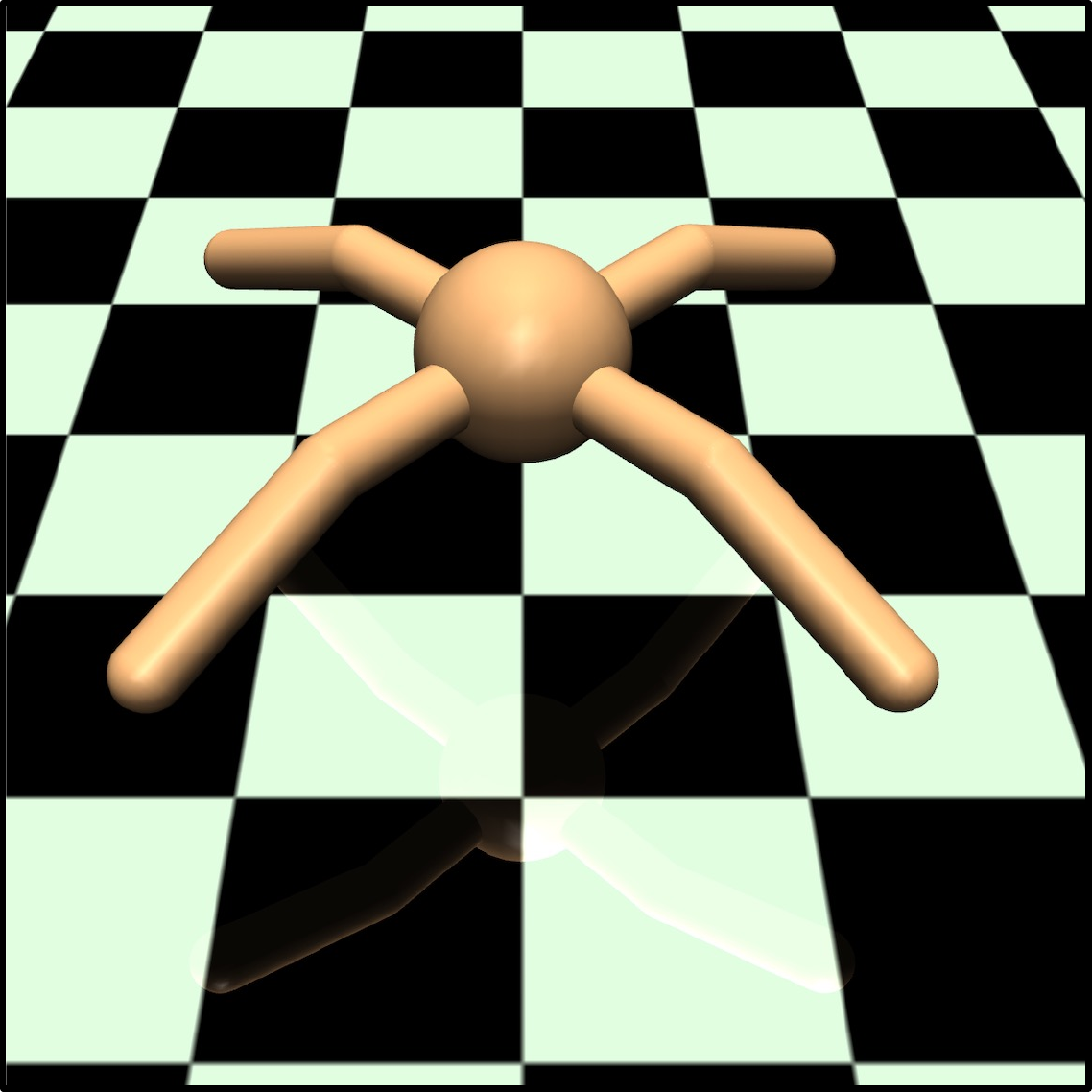}}\hspace{1em}
\subfigure[]{\includegraphics[width=0.4\columnwidth]{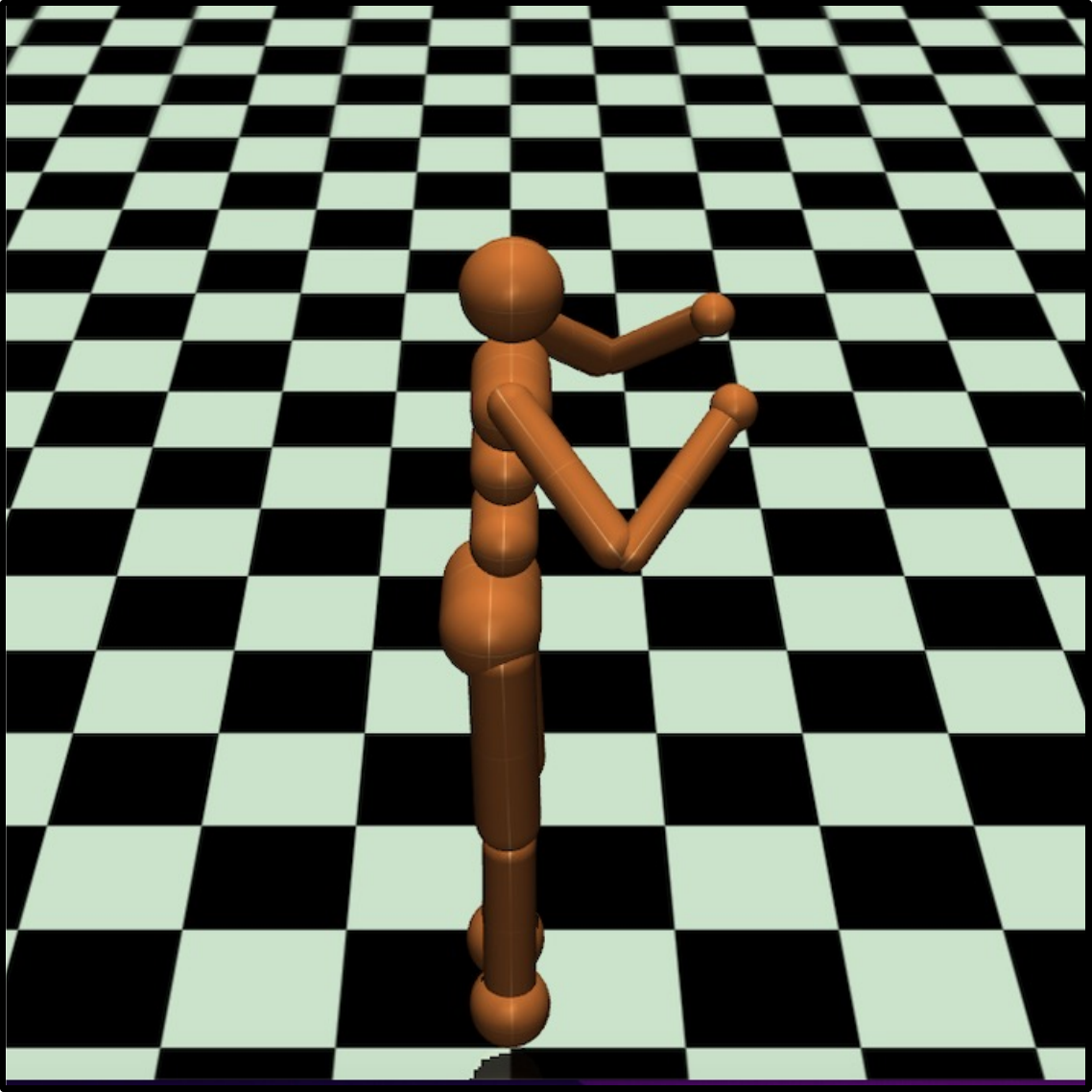}}
 \caption{Two types of dynamic environments. (a) Ant-Navigation domain. (b) Humanoid-Navigation domain.}
 \label{fig2-env} 
\end{figure}

In the experiment of this section, we choose two high-dimensional dynamic environments from MuJoCo~\cite{todorov2012mujoco}, as shown in Fig.~\ref{fig2-env}, and the details are as follows.

\subsubsection{Ant-Navigation}
The first experiment consists of a variation of the Ant-v3, which makes an ant agent reach the specified 2-D positions.
The states are 113-dimensional vectors, and the actions are 8-dimensional vectors clipped to be in the range of $[-1, +1]$.
In our experiment, the ant agent only needs to be within $0.2$ of the goal from the starting point or reach the maximum number of steps $M = 100$ regardless of speed. 
The reward consists of three parts: the negative Euclidean distance to the goal, a controlled cost that is positively related to the scale of actions, and a contact cost that is positively related to the scale of contact forces. 
We set up four source tasks with different goal positions and gears on the legs of the ant agent, where the gear is used to specify 3D force and torque axes by scaling the length of the actuator. The goal positions are $(1,1)$, $(-1,1)$, $(-1,-1)$, $(1,-1)$, and the corresponding gears are $50$, $100$, $150$, $200$, respectively.   
When faced with target tasks similar to the source ones, we need to select the most appropriate policy from the library. 
To solve this problem, the tuple $\left(s, a, r, s'\right)$ is used as the observation signal.

\subsubsection{Humanoid-Navigation}
We design the fifth experiment to verify that our method can solve very high-dimensional problems. We adopt a variation of the Humanoid-v3, which makes a humanoid agent reach the specified 2-D position. It is very challenging to solve this high-dimensional problem with a continuous state-action space. The states are 378-dimensional vectors, and the actions are 17-dimensional vectors clipped to be in the range of $[-0.4, +0.4]$.
In this experiment, the humanoid agent will end the episode when it reaches the specified range of the goal or the maximum number of steps $M=1000$, or it falls.
The reward contains four parts: the negative Euclidean distance to the goal, an alive bonus when the $z$-coordinate of the agent is in the specified range, a large bonus when the agent reaches its goal, and a control and impact cost.
We set up four source tasks where the goal positions are $(0.6, 0.6)$, $(-0.55, 0.55)$, $(0.5, -0.5)$, and $(-0.45, -0.45)$, and the corresponding specified ranges are $0.4$, $0.35$, $0.3$, $0.25$, respectively.
To solve this problem, we use the tuple $\left(s, a, r\right)$ as the observation signal. 

Analogous to Section~\ref{GP-experiment},
we select some target tasks that are relatively close to the source tasks in each domain.
\begin{itemize}
    \item In Ant-Navigation domain, we select twelve target tasks, and their goal positions and gears are ($1.1, 1.1, 60$), ($0.9, 0.9, 40$), ($1.1, -1.1, 90$), ($0.9, -0.9, 110$), ($-1.1, \\1.1, 160$), ($-0.9, 0.9, 140$), ($-1.1, -1.1, 190$), ($-0.9, \\-0.9, 210$), ($1, 1, 50$), ($1, -1, 100$), ($-1, 1, 150$), ($-1, \\-1, 200$).
    \item In Humanoid-Navigation domain, we select twelve target tasks, and their goal positions and specified ranges are ($0.55, 0.55, 0.4$), ($0.65, 0.65, 0.4$), ($-0.5, 0.5, 0.35$), ($-0.6, 0.6, 0.35$), ($0.45, -0.45, 0.3$), ($0.55, -0.55, 0.35$), \\($-0.5, -0.5, 0.25$), ($-0.4, -0.4, 0.25$), ($0.6, 0.6, 0.4$), \\($-0.55, 0.55, 0.35$), ($0.5, -0.5, 0.3$),  ($-0.45, -0.45, \\0.25$).
\end{itemize}

\begin{figure*}[tb]
\centering
\subfigure[]{\includegraphics[width=0.32\textwidth]{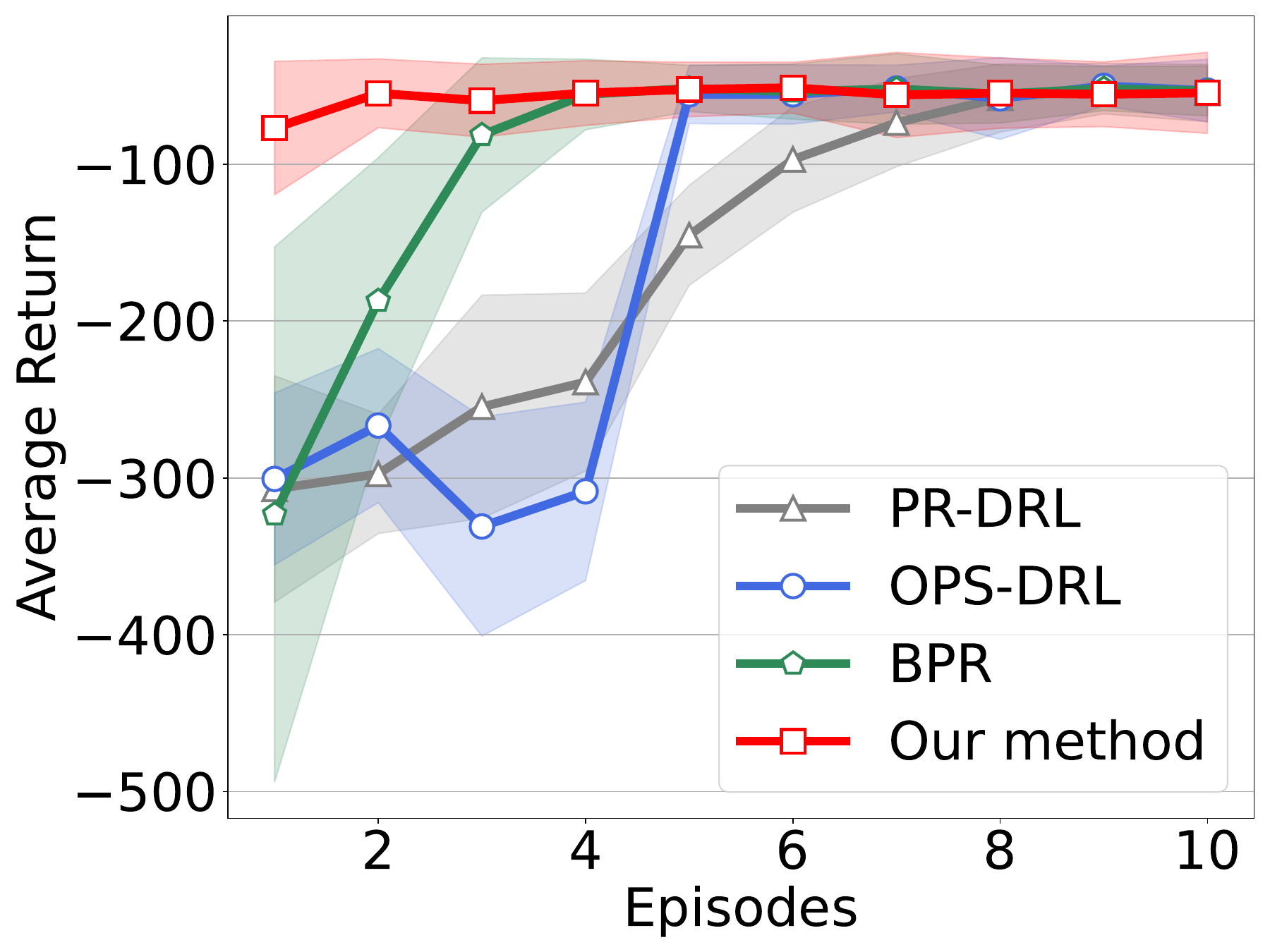}}\hspace{2em}
\subfigure[]{\includegraphics[width=0.32\textwidth]{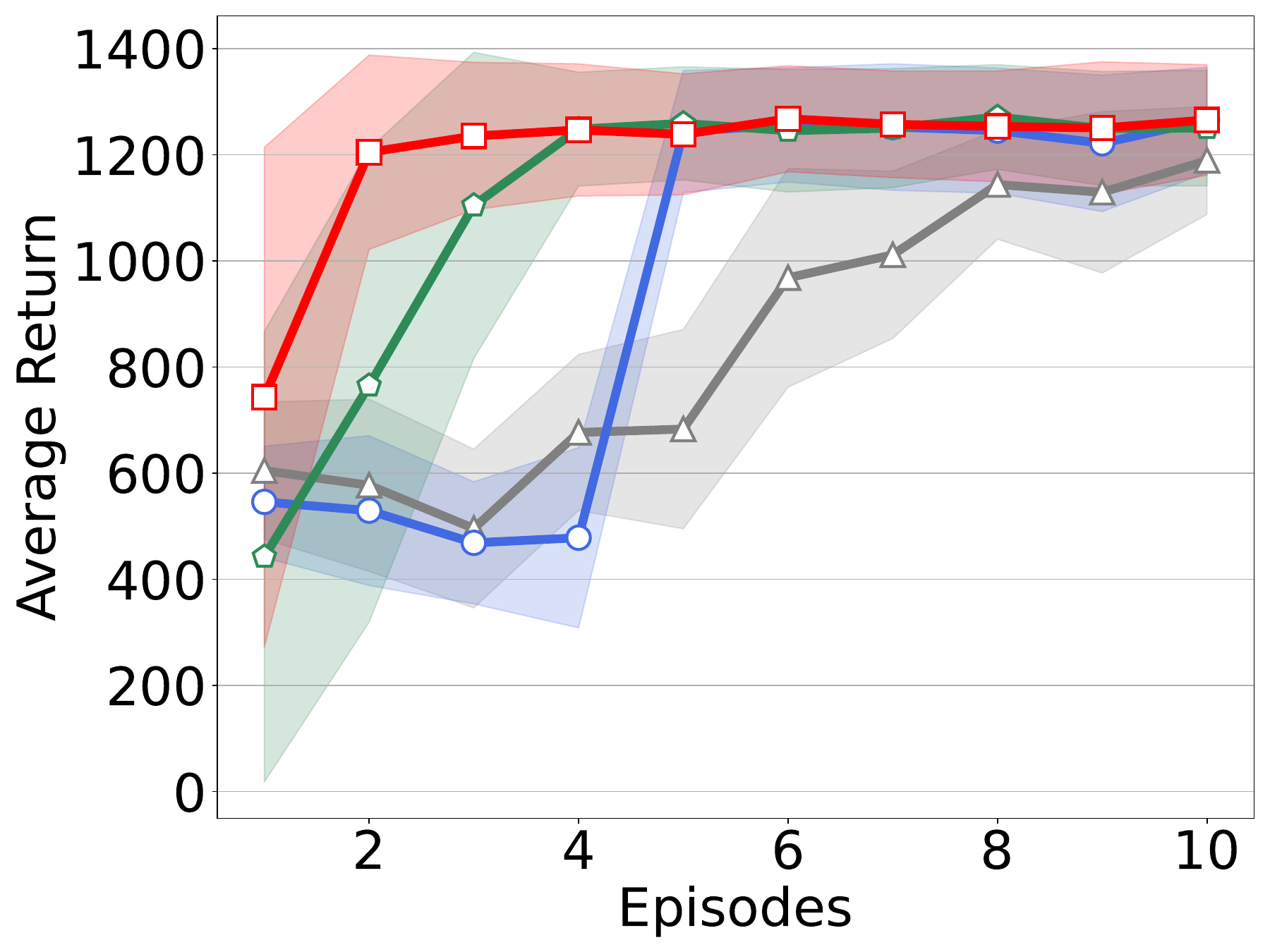}}
\caption{Average return per episode of all tested methods in: (a) Ant-Navigation. (b) Humanoid-Navigation. }
\label{fig3}
\end{figure*}

\begin{table*}[tb]
\centering
\caption{Average return over all episodes of all test methods implemented in two domains.}
\renewcommand\arraystretch{1.2}
\begin{tabular}{c|c|c|c|c}
\cmidrule[\heavyrulewidth]{1-5}
Method & PR-DRL & OPS-DRL &  BPR & Our method\\
\hline

Ant-Navigation  & $-157.79\pm 109.36$       & $-153.05\pm 128.53$ &  $-96.46 \pm 107.31$    & $\mathbf{-56.94\pm 25.78}$ \\

Humanoid-Navigation  & $848.15 \pm 295.14$       & $950.91 \pm 384.62$ &  $1108.84 \pm 354.02$    & $\mathbf{1196.59\pm 243.20}$ \\

\cmidrule[\heavyrulewidth]{1-5}
\end{tabular}
\label{tab3}
\end{table*}

\begin{figure*}[tb]
\centering
\subfigure[]{\includegraphics[width=0.32\textwidth]{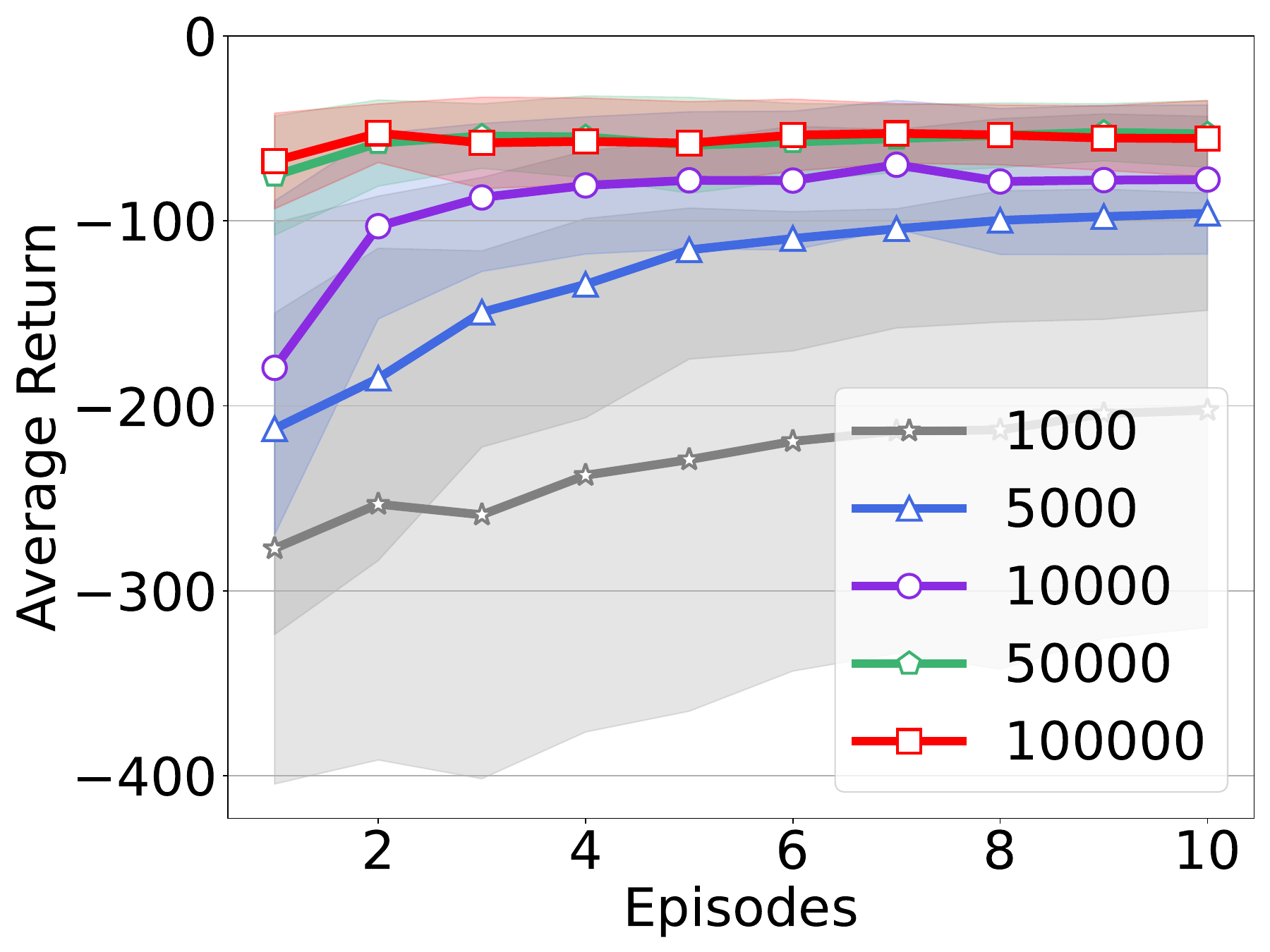}}\hspace{2em}
\subfigure[]{\includegraphics[width=0.32\textwidth]{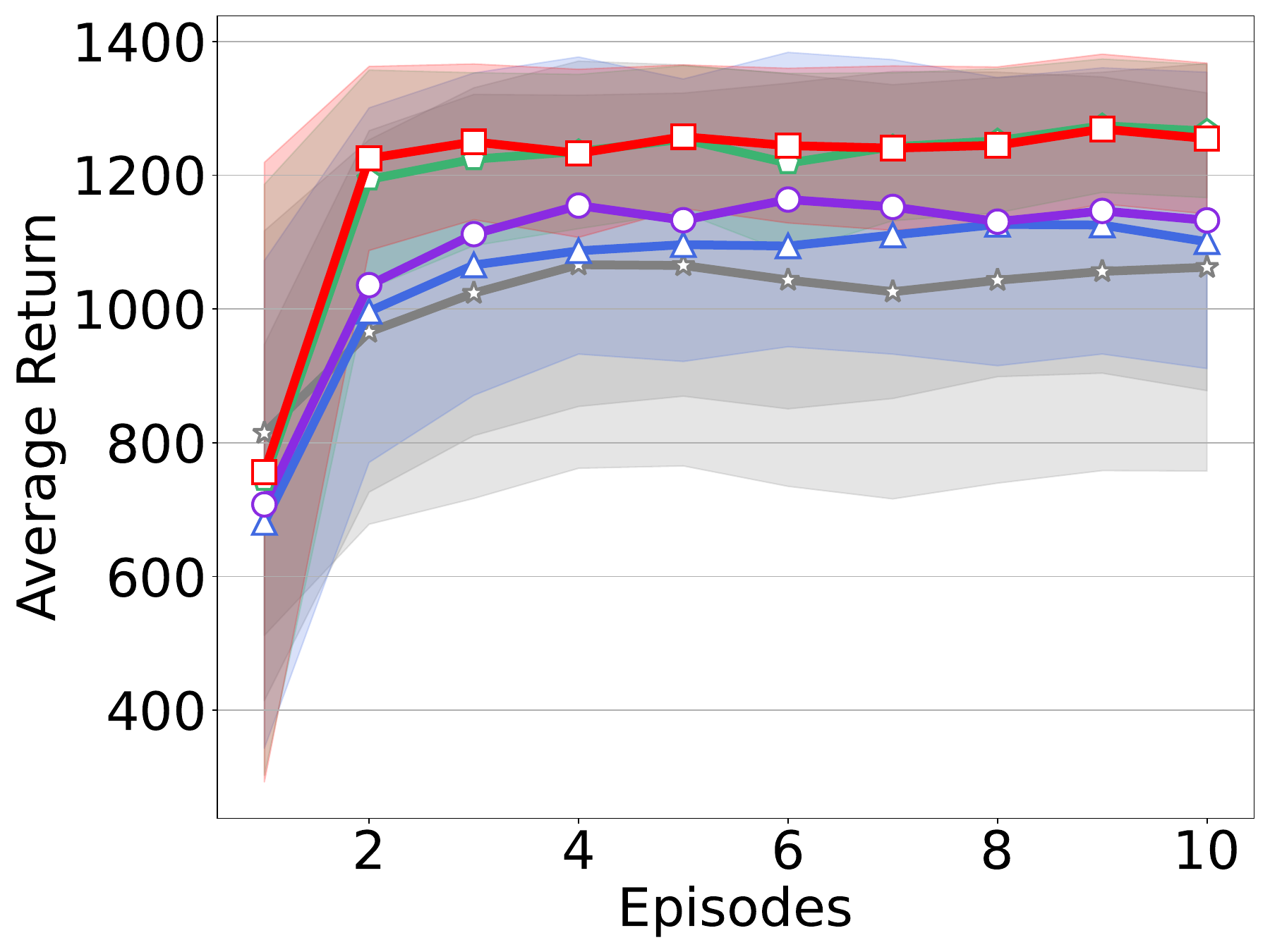}}
 \caption{Average return per episode of different sample sizes used for NNs in: (a) Ant-Navigation; (b) Humanoid-Navigation. }
\label{fig4}
\end{figure*}

\begin{table*}[htb]
\centering
\caption{Average return over all episodes of different sample sizes used for NNs in two domains.}
\renewcommand\arraystretch{1.2}
\begin{tabular}{c|c|c|c|c|c}
\cmidrule[\heavyrulewidth]{1-6}
Sample size & 1000 & 5000 &  10000 & 50000 & 100000\\
\hline

Ant-Navigation  & $-230.84\pm 132.05$      & $-130.40\pm 81.34$ &  $-91.06 \pm 56.40$  &  $-57.28\pm 22.61$   & $\mathbf{-56.34\pm 20.95}$ \\

Humanoid-Navigation & $1016.49 \pm 311.26$  & $1048.26 \pm 273.24$       & $1086.82 \pm 276.53$ &  $1190.30\pm 234.84$    & $\mathbf{1197.40\pm 236.57}$ \\

\cmidrule[\heavyrulewidth]{1-6}
\end{tabular}
\label{tab4}
\end{table*}

In this experiment, the number of learning episodes in the target task, $K$, is set as 10.
The hyper-parameters are set as:  $\varepsilon^2_{NN} = 0.1$ for the Ant-Navigation domain and $\varepsilon^2_{NN} = 1$ for the   Humanoid-Navigation domain.
To solve complex problems, we use the twin delayed deep deterministic policy gradient (TD3) algorithm~\cite{fujimoto2018addressing} to learn optimal policies for source tasks.

We present the primary results of our NN-based method and all baseline methods on two complex experimental domains, and all the results are averaged over multiple target tasks.
Fig.~\ref{fig3} shows the average return per learning episode, and Table~\ref{tab3} reports the numerical results in terms of average return over all learning episodes.
BPR obtains better policy transfer performance than PR-DRL and OPS-DRL, and our method achieves much more efficient policy transfer in all tasks compared to BPR. 
In domains with high-dimensional, our method can also achieve much better jump-start performance compared to all baselines, as illustrated in Figs.~\ref{fig3}-(a) and~\ref{fig3}-(b). 
Further, from the numerical results in Table~\ref{tab3}, our method generally obtains the largest average return over all learning episodes in all domains.
In addition, it can be observed from the statistical results that our method mostly obtains smaller confidence intervals and standard errors than the baselines.
This phenomenon indicates that our method can provide more stable source task selection and knowledge transfer in high-dimensional domains.

Moreover, we study how the number of state transition samples $(s,a,r,s')$ used for fitting the NNs affects the performance of our method.
We use $1000$, $5000$, $10000$, $50000$, and $100000$ samples, respectively, for fitting the state transition functions using NNs in all domains to observe the performance of our method.
Fig.~\ref{fig4} shows the average return per learning episode, and Table~\ref{tab4} reports the numerical results in terms of average return over all learning episodes.
In all domains, as the sample size continues to grow, the performance of our method is improved, but the degree of performance improvement is decreasing.
When the sample size reaches $50000$, the performance is almost optimal.
It demonstrates that using the limited number of samples to fit the state transition function, our method based on NN is capable of accurately estimating the scalable observation model and inferring the task belief for efficient policy transfer.

Overall, it is verified that our method achieves a more accurate inference of the task belief and converges more quickly to the most appropriate policy for the target task.
Using the scalable observation model with more informative observation signals, our method can efficiently update the task belief and achieve better performance compared to all baselines in all experimental domains.

\subsection{Results for Continual Learning}\label{Continual-experiment}
In this experiment, we evaluate our method and the baseline approach in continual learning settings where the agent is faced with a new unknown target task that largely differs from any of the source tasks.
\begin{itemize}
    \item In 2-D Navigation domain, we select four target tasks, and their goal positions are ($0, 10$), ($0, -9$), ($-8, 0$), ($9, 0$).
    \item In CartPole domain, we select two target tasks, and their values of the additional force $F^{\prime}$ are $8$, $-8$.
    \item In LunarLander domain, we select two target tasks, and their value of additive powers are ($0.5, 0.5$), ($-0.5, 0.5$).
    \item In Ant-Navigation domain, we select four target tasks, and their goal positions and gears are ($1,0,50$), ($-0.9,0,100$), ($0,-0.8,150$), ($0,0.9,200$).
    \item In Humanoid-Navigation domain, we select four target tasks, and their goal positions and specified ranges are ($0.2,0.7,0.3$), ($-0.65,0.2,0.3$),
    ($0.25,-0.6,0.3$), ($0,-0.8,0.3$).
\end{itemize}

Different from the previous settings, our method is required to learn a new policy for the target task in an online fashion, other than selecting an existing source policy from the offline library, when a sufficiently different target task is detected.
In this setting, we compare our method to the BPR and we deploy the agents to some target tasks that are not close to the source ones, and the final received returns are shown in Table~\ref{tab5}.
It is observed that BPR performs worse in the five domains since it is unable to select an appropriate policy from the offline library to respond to a new unknown task.
In contrast, our method can learn a new optimal policy for the target task by expanding the source library in a continual learning manner.
In the policy reuse phase, our method can detect the unknown tasks according to the received return, and switch to the learning phase to learn a new optimal policy for the target task, thus effectively avoiding negative transfer.

\begin{table*}[htb]
\centering
\renewcommand\arraystretch{1.2}
\caption{The final received returns implemented in all domains with continual learning settings.}
\begin{tabular}{c|c|c|c|c|c}
\cmidrule[\heavyrulewidth]{1-6}
Domain & 2-D Navigation & CartPole & LunarLander & Ant-Navigation & Humanoid-Navigation\\
\hline
%
%

BPR         & $-895.15\pm 292.44$      & $34.0  \pm 10.34$ &  $-33.25\pm 87.71$    & $-288.05\pm 62.82$  & $ 343.61\pm 110.79$\\

Our method         & $\mathbf{-39.82\pm 5.39}$               & $\mathbf{96.44\pm 2.90}$ &  $\mathbf{-24.96\pm 5.29}$    & $\mathbf{-70.08 \pm 11.30}$  & $\mathbf{1347.82 \pm 89.98}$ \\

\cmidrule[\heavyrulewidth]{1-6}
\end{tabular}
\label{tab5}
\end{table*}

\section{Conclusions and Future Work}\label{conclusions}
In the paper, we proposed a general improved BPR framework to implement more efficient policy transfer in DRL, which can be implemented by any DRL algorithm, whether they are model-based or model-free, on-policy or off-policy.
We introduced a scalable observation model by using the non-parametric GP or parametric NN to fit the state transition function of source tasks in a model-based way, which achieves more efficient policy reuse with informative and instantaneous observation signals. 
GP is merited for its sample efficiency and the ability to provide uncertainty measurements on the predictions, while NN is preferred for extremely high-dimensional tasks.
Moreover, we extended our method to continual learning settings conveniently in a plug-and-play fashion to avoid negative transfer.

While we use GP and NN to estimate the observation model, our method is a general framework and can be easily combined with any distribution matching technique or probabilistic model. 
Thus, a potential direction for future work is to use more powerful techniques to estimate the observation model, such as more advanced GPs that can scale to high-dimensional domains~\cite{csato1999efficient, csato2002sparse, lawrence2002fast, tipping2001sparse}, or Bayesian neural networks.
In addition, we can also consider taking different kernel functions for states and actions according to specific situations.
Alternatively, a potential solution is to use the return distribution of distributional RL approaches to avoid introducing any additional models.
Another direction is to improve the settings of continual learning, for example, using policy distillation to speed up the learning process for unknown tasks.
Furthermore, another significant and challenging direction is to provide some theoretical guarantees for the family of BPR algorithms such as the quantitative analysis of sample efficiency.

\appendix[Baselines: BPR, PR-DRL, and OPS-DRL]
\subsection{BPR}
The family of BRP algorithms~\cite{rosman2016bayesian,hernandez2016bayesian,yang2018towards,zheng18deep,zheng2021efficient,gao2022bayesian} typically uses the episodic return as the observation signal, and needs to apply all source policies on all source tasks to estimate the tabular-based observation model in an
offline manner. 
However, it requires a large amount of episode return samples to estimate the probability distribution of the observation model. 
In our experiments, for better applicability, we employ a variant of the BPR algorithms, which models the probability distribution of the observation model as a Gaussian distribution. 
For each task-policy pair $(\tau,\pi)$, we apply the source policy $\pi$ on the source $\tau$, and repeat it one hundred times. 
Then, we take the mean of episode returns $\mu_{U}$, and artificially choose an appropriate variance $\varepsilon^2_{U}$. 
In this way, we can obtain the observation model as $\mathrm{P}(\sigma \mid \tau, \pi) \sim \mathcal{N}\left(\mu_{U}, \varepsilon^2_{U}\right)$.

\subsection{PR-DRL}
The probabilistic policy reuse algorithm~\cite{fernandez2010probabilistic} improves its exploration by exploiting the source policies probabilistically. 
It updates the probability depending on the reuse gains, which are obtained concurrently during the learning process. 
The original probabilistic Policy Reuse (PR) algorithm was implemented by the Q-learning, referred to as PRQ-learning, which can be applied to simple domains with a discrete state-action space only.
Here, we adopt a version of PRQ-learning that builds on DRL, similar to the experimental settings in~\cite{barreto2016successor,wang2019tnnls}.
Thus, the resulting approach is called PR-DRL which is directly comparable to our method.
Algorithm~\ref{alg2} shows its pseudocode, where the initial value of the temperature parameter $\nu$ is $0$, and the value of the incremental size $\Delta \nu$ is $0.05$.

\subsection{OPS-DRL}
The Optimal Policy Selection (OPS) algorithm~\cite{li2018optimal} formulates online source policy selection as a multi-armed bandit problem and augments Q-learning with policy reuse. 
Analogous to the setting of PR-DRL, we adopt a version of OPS that builds on DRL and obtain another baseline approach OPS-DRL.
Note that the original OPS algorithm initializes the reuse gains of source policies by applying all the source policies on the target task.
For a fair comparison, we set the initial reuse gains of source policies in the OPS-DRL as zeros.
The pseudocode is shown in Algorithm~\ref{alg3}.
Moreover, PR-DRL and OPS-DRL learn a new policy for the target task while reusing source policies from the policy library. 
For a fair comparison with our method, we assume that the two baselines only reuse source policies without learning a new one. 

\begin{algorithm}[tb]
\caption{PR-DRL}
\label{alg2}
\KwIn{
      Source policy library $\{\pi_j\}_{j=1}^n$, \newline 
      number of episodes $K$, \newline 
      the maximum number of steps $M$.
      }

${W}_{j}$ $\leftarrow$0, the reuse gain associated with $\pi_{j}$,\\
${V}_{j}$ $\leftarrow$ 0, the number of times $\pi_{j}$ is used.\\
Temperature parameter $\nu$ and the incremental size $\Delta \nu$ \\

\While{episode $\leq$ $K$}{  

    \While{$j$ $\leq$ $n$}{
      $p_{j} \leftarrow e^{v \times \operatorname{W}_{j}} / \sum^n_{j'} e^{v \times \operatorname{W}_{j'}}$}
    Select the reuse policy $\pi_{j}$ according $P$.\\
    
    $U \leftarrow 0$\\
    \While{step $\leq$ $M$  $\&$ not reaching the goal}{
        Apply $\pi_{j}$  on target task $\tau_{0}$ and receive the reward $r_{t}$.\\
        $U = U + \gamma^{t} r_{t} $
    }
    
    $W_{j} \leftarrow \frac{W_j \times V_j+ U}{V_j+1}$\\
    $V_{j} \leftarrow$ $V_{j}+1$ \\
    $\nu = \nu +\Delta \nu$\\

}
\end{algorithm}
\begin{algorithm}[tb]
\caption{OPS-DRL}
\label{alg3}
\KwIn{
      Source policy library $\{\pi_j\}_{j=1}^n$, \newline 
      number of episodes $K$, \newline 
      the maximum number of steps $M$.
      }
${W}_{j}$ $\leftarrow$0, the reuse gain associated with $\pi_{j}$,\\
${V}_{j}$ $\leftarrow$ 0, the number of times $\pi_{j}$ is used.\\

\While{episode $\leq$ $K$}{  

    Select the reuse policy $\pi_{j}$ according:  $j=\underset{1 \leq j \leq n}{\arg \max }\left({W}_{j}+\sqrt{\frac{2 \ln \left(\sum_{j=1}^{n} {V}_{j}+1\right)}{{V}_{j}+1}}\right)$ .\\
    
    $U \leftarrow 0$\\
    \While{step $\leq$ $M$  $\&$ not reaching the goal}{
        Apply $\pi_{j}$  on target task $\tau_{0}$ and receive the reward $r_{t}$.\\
        $U = U + \gamma^{t} r_{t} $
    }
    
    $W_{j} \leftarrow \frac{\text {W}_{j} \times \text {V}_{j}+ U}{\text {V}_{j}+1}$\\
    $V_{j} \leftarrow$ $V_{j}+1$ \\

}
\end{algorithm}

\footnotesize
\bibliography{Tnnls}
\bibliographystyle{IEEEtran}

\begin{IEEEbiography}[{\includegraphics[width=1.0in,height=1.25in,clip,keepaspectratio]{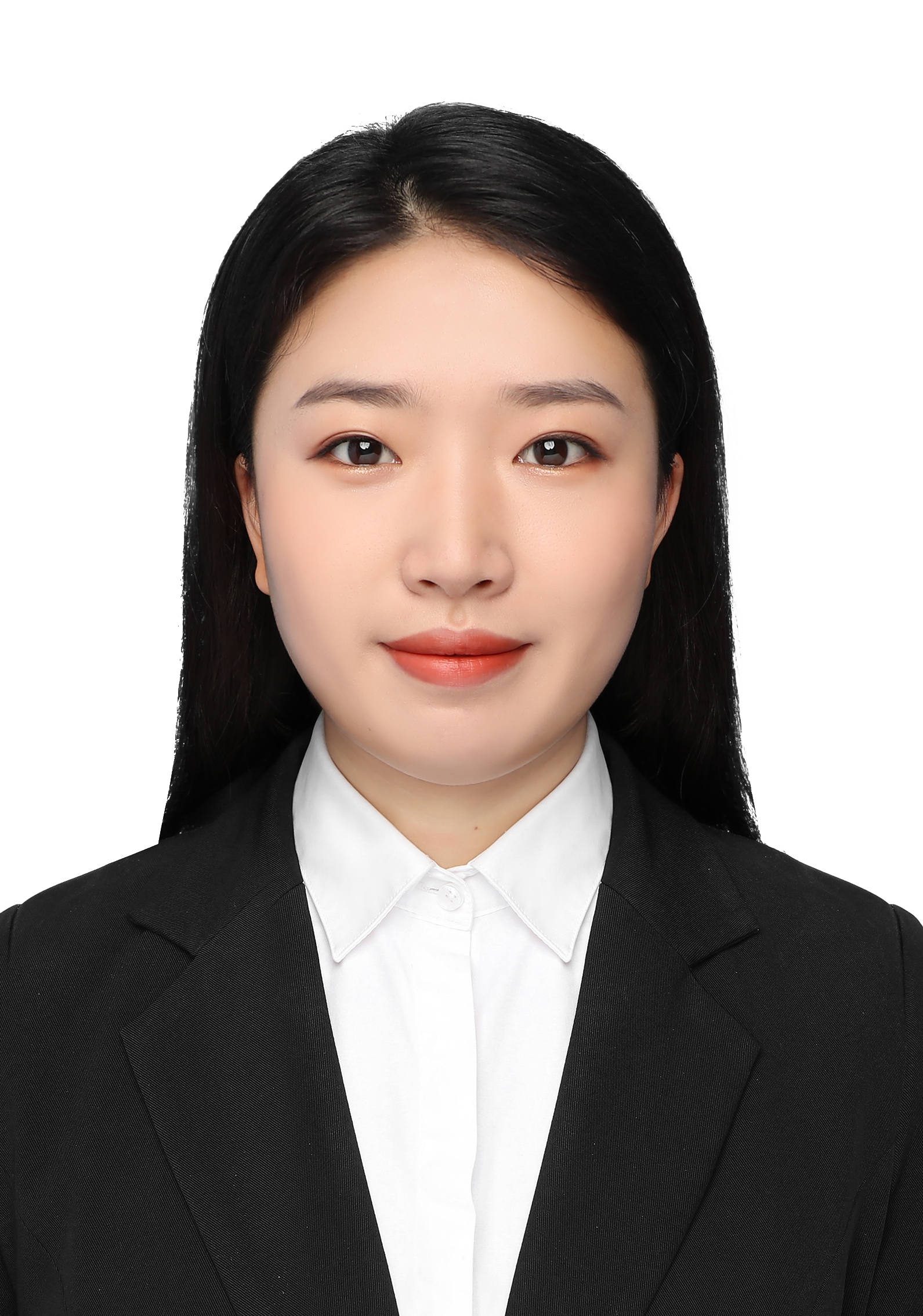}}]{Jinmei Liu} (S'20) is pursuing the Ph.D. degree at the Department of Control Science and Intelligent Engineering, School of Management and Engineering, Nanjing University, Nanjing, China, where she obtained the B.E. degree in automation and the M.E. degree in control science and intelligent engineering in 2020 and 2023, respectively.
Her research interests include reinforcement learning and machine learning.

\end{IEEEbiography}

\begin{IEEEbiography}[{\includegraphics[width=1.0in,height=1.25in,clip,keepaspectratio]{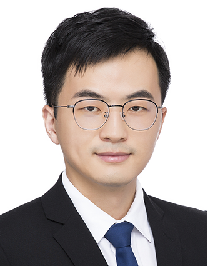}}]{Zhi Wang} (S'19-M'20) received the Ph.D. degree in machine learning from the Department of Systems Engineering and Engineering Management, City University of Hong Kong, Hong Kong, China, in 2019, and the B.E. degree in automation from Nanjing University, Nanjing, China, in 2015. He is currently an Associate Professor with the Department of Control Science and Intelligent Engineering, School of Management and Engineering, Nanjing University, Nanjing, China. He holds visiting positions at the University of New South Wales, Australia and the State Key Laboratory of Management and Control for Complex Systems, Institute of Automation, Chinese Academy of Sciences, China.
 
His current research interests include reinforcement learning, machine learning, and robotics.
He served as the Associate Editor for IEEE International Conference on Systems, Man, and Cybernetics 2021, 2022, and 2023, and the Associate Editor for IEEE International Conference on Networking, Sensing, and Control 2020.
\end{IEEEbiography}

\begin{IEEEbiography}[{\includegraphics[width=1.0in,height=1.25in,clip,keepaspectratio]{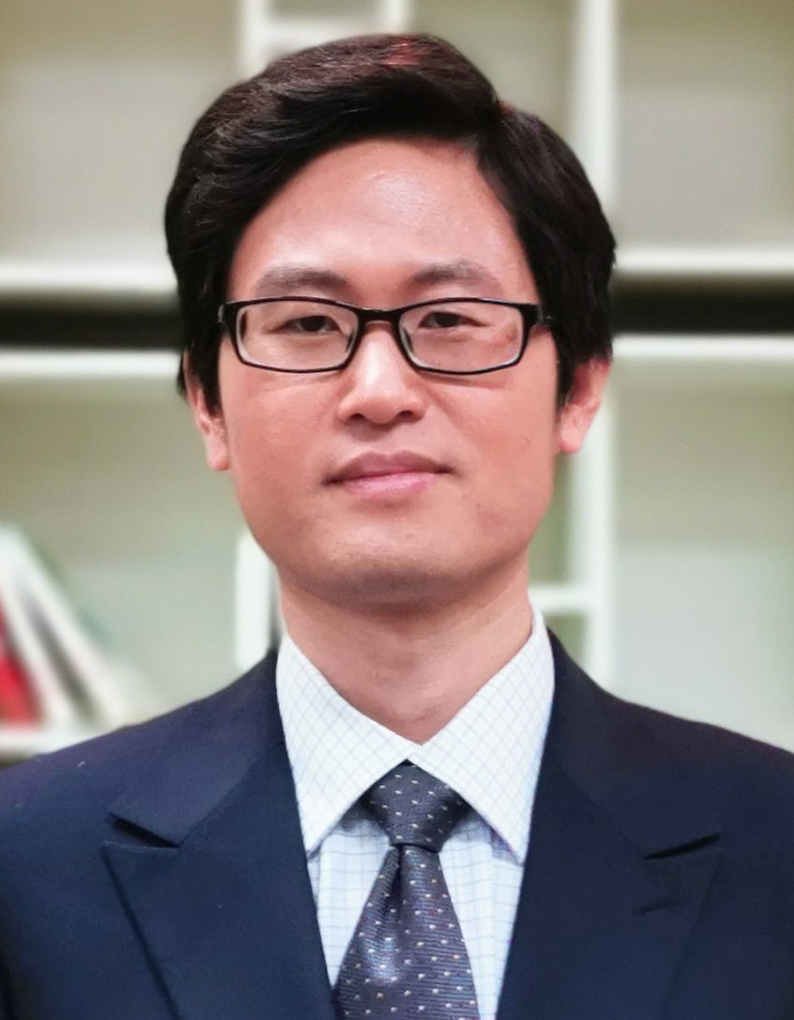}}]{Chunlin Chen}
(S'05-M'06-SM'21) received the B.E. degree in automatic control and Ph.D. degree in control science and engineering from the University of Science and Technology of China, Hefei, China, in 2001 and 2006, respectively. 
He is currently a full professor and the vice dean of School of Management and Engineering, Nanjing University, Nanjing, China. 
He was a visiting scholar at Princeton University, Princeton, USA, from 2012 to 2013. He had visiting positions at the University of New South Wales, Canberra, Australia, and the City University of Hong Kong, Hong Kong, China.

His recent research interests include reinforcement learning, mobile robotics, and quantum control. 
He is the Chair of Technical Committee on Quantum Cybernetics, IEEE Systems, Man and Cybernetics Society.
\end{IEEEbiography}

\begin{IEEEbiography}[{\includegraphics[width=1.0in,height=1.25in,clip,keepaspectratio]{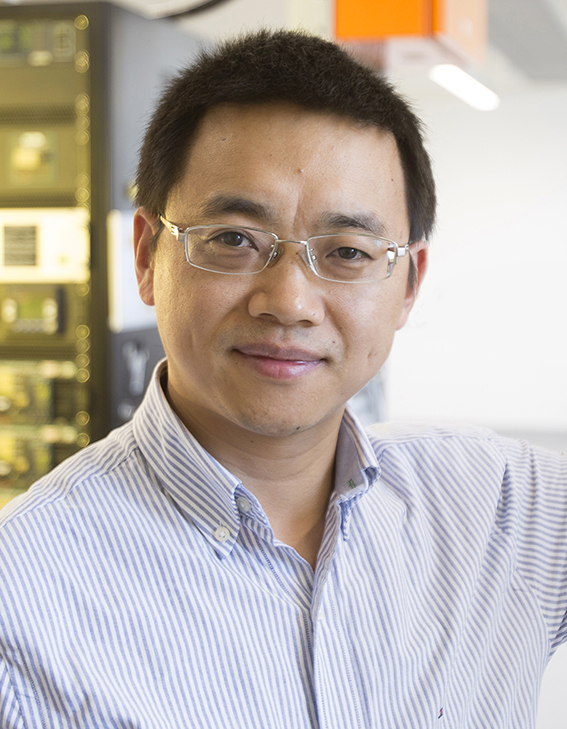}}]{Daoyi Dong}
received the B.E. degree in automatic control and the Ph.D. degree in engineering from the University of Science and Technology of China, Hefei, China, in 2001 and 2006, respectively.

He was an Alexander von Humboldt Fellow at AKS, University of Duisburg-Essen, Duisburg, Germany.
He was with the Institute of Systems Science, Chinese Academy of Sciences, Beijing, China, and with Zhejiang University, Hangzhou, China. He had visiting positions at Princeton University, NJ, USA; RIKEN, Wako-Shi, Japan; and The University of Hong Kong, Hong Kong. 
He is currently an Associate Professor at the University of New South Wales, Canberra, ACT, Australia. His research interests include quantum control and machine learning.

Dr. Dong was awarded the ACA Temasek Young Educator Award by the Asian Control Association and was a recipient of Future Fellowship, the International Collaboration Award and the Australian Post-Doctoral Fellowship from the Australian Research Council, and a Humboldt Research Fellowship from the Alexander von Humboldt Foundation of Germany. 
He was a Member-at-Large, Board of Governors, and the Associate Vice President for Conferences and Meetings, IEEE Systems, Man and Cybernetics Society. He served as an Associate Editor for the IEEE TRANSACTIONSON NEURAL NETWORKS AND LEARNING SYSTEMS from 2015 to 2021. He is currently an Associate Editor of the IEEE TRANSACTIONS ON CYBERNETICS and a Technical Editor of the IEEE/ASME TRANSACTIONS ON MECHATRONICS. 
He is a Fellow of the IEEE.
\end{IEEEbiography}

\end{document}